\documentclass[journal]{IEEEtran}
\usepackage{cite}
\usepackage{amsmath,amssymb,amsfonts}
\usepackage{algorithmic}
\usepackage{graphicx}
\usepackage{textcomp}
\usepackage{xcolor}
\usepackage[ruled,linesnumbered]{algorithm2e}
\usepackage{multirow}
\usepackage{enumerate}
\usepackage{url}
\usepackage{booktabs}
\usepackage{threeparttable}
\usepackage{endnotes}
\usepackage{subfigure}
\usepackage{comment}
\usepackage[switch,pagewise,columnwise]{lineno}
\ifCLASSINFOpdf
\else
\fi

\begin{document}
%
%\linenumbers

\title{Variable Population Memetic Search: A Case Study on the Critical Node Problem}
\author{Yangming~Zhou,
        Jin-Kao~Hao,
        Zhang-Hua~Fu,
        Zhe~Wang,
        and Xiangjing Lai
\thanks{This work was partially supported by the National Natural Science Foundation of China under Grant 61903144, the Shanghai Sailing Program under Grant 19YF1412400, the Fundamental Research Funds for the Central Universities of China under Grant 222201817006, and the Shenzhen Science and Technology Innovation Commission under Grant JCYJ20180508162601910. (Corresponding authors: Jin-Kao Hao and Zhang-Hua Fu.)}
\thanks{Y.~Zhou and Z.~Wang are with the School of Information Science and Engineering, East China University of Science and Technology, 130 Meilong Road, 200237 Shanghai, China. Y. Zhou is also with the Key Laboratory of Advanced Control and Optimization for Chemical Processes, Ministry of Education (e-mails: ymzhou@ecust.edu.cn, wangzhe@ecust.edu.cn).}
\thanks{J.-K.~Hao is with the Department of Computer Science, LERIA, Universit\'{e} d'Angers, 2 Boulevard Lavoisier, 49045 Angers, France and the Institut Universitaire de France, 1 rue Descartes, 75231 Paris, France (e-mail: jin-hao.hao@univ-angers.fr).}
\thanks{Z.~Fu is with the Robotics Laboratory for Logistics Service, Institute of Robotics and Intelligent Manufacturing, The Chinese University of Hong Kong and the Shenzhen Institute of Artificial Intelligence and Robotics for Society, 518172 Shenzhen, China (e-mail: fuzhanghua@cuhk.edu.cn).}
\thanks{X.~Lai is with the Institute of Advanced Technology, Nanjing University of Posts and Telecommunications, 210023 Nanjing, China (e-mail: laixiangjing@gmail.com).}
%\thanks{* Corresponding authors.}
}

%\markboth{IEEE TRANSACTIONS ON }%

\maketitle

\begin{abstract}

Population-based memetic algorithms have been successfully applied to solve many difficult combinatorial problems. Often, a population of fixed size was used in such algorithms to record some best solutions sampled during the search. However, given the particular features of the problem instance under consideration, a population of variable size would be more suitable to ensure the best search performance possible. In this work, we propose variable population memetic search (VPMS), where a strategic population sizing mechanism is used to dynamically adjust the population size during the memetic search process. Our VPMS approach starts its search from a small population of only two solutions to focus on exploitation, and then adapts the population size according to the search status to continuously influence the balancing between exploitation and exploration. We illustrate an application of the VPMS approach to solve the challenging critical node problem (CNP). We show that the VPMS algorithm integrating a variable population, an effective local optimization procedure (called diversified late acceptance search) and a backbone-based crossover operator performs very well compared to state-of-the-art CNP algorithms. The algorithm is able to discover new upper bounds for 13 instances out of the 42 popular benchmark instances, while matching 23 previous best-known upper bounds.

\end{abstract}


% \fi
\IEEEpeerreviewmaketitle

\section{Introduction}
\label{Sec:Introduction}

Memetic algorithms (MAs) are hybrid metaheuristics that combines local optimization and evolutionary search \cite{Moscato&Cotta2003}. By hybridizing these two different search methods, MAs are expected to benefit from their complementary search strategies. Since their introduction, MAs have been applied with success to numerous search problems including popular NP-hard problems (e.g., graph coloring \cite{Porumbeletal2010}, maximum diversity \cite{Zhou2017a}, and quadratic assignment \cite{Benlic2015}) and practical applications (e.g., identification of critical nodes in sparse graphs \cite{Zhou2019}, influence maximization in multiplex networks \cite{Wang2019}, vehicle routing \cite{Feng2014}). Comprehensive surveys of recent research in memetic computation and additional application examples can be found in, e.g.,  \cite{Chen2011,Neri2012}.

To design an effective memetic algorithm for a given problem, one needs to specify a number of algorithmic components including the local optimization procedure, the crossover operator, and the pool updating strategy \cite{Krasnogor2005}. Additionally, since MAs rely on a population of individuals, the population size needs to be identified as well. Our literature review indicates that existing studies on MA applications focus mainly on designing algorithmic components such as local optimization and crossover, while the population size is typically fixed to a constant value which is kept unchanged during the search.

Meanwhile, it is known that for a memetic algorithm, the population size impacts its solution quality and the running time \cite{Hart2004}. Indeed, there is a general consensus that a small population implies a low population diversity and may lead to premature convergence of the algorithm, whereas a large population promotes diversity, nevertheless consumes more computational resources. Moreover, for a population-based evolutionary algorithm, it is recognized that the optimal population size depends on the problem instance under consideration \cite{Eiben2004} and can even vary at different evolution stages of the algorithm \cite{Weise2016}. Specifically, for MAs in discrete optimization, the importance of selecting a proper population size was investigated in the context of solving a particular assignment problem \cite{Karapetyan2011} and to our knowledge, this is the only study in the literature dedicated to population sizing for MAs applied to combinatorial problem.

In this work, we present variable population memetic search (VPMS) where a strategic population sizing mechanism is integrated to the memetic computation framework. This work was motivated by the importance of MAs and the scarcity of research on dynamic population sizing in MAs. We summarize the contributions of the work as follows.

First, from an algorithmic perspective, the proposed variable population memetic search enhances the memetic computation framework with a strategic population sizing mechanism to dynamically influence the balancing between exploration and exploitation. A VPMS algorithm starts its search with a small population of two individuals (solutions) to favor exploitation. Upon reaching local optima solutions, the population is augmented with new high-quality solutions to strengthen population diversity and enhance exploration of the search space. When the population reaches a maximum allowable size while the search is still stagnating, it is shrunk to two individuals while maintaining the best solution found so far to start a new round of exploitation and exploration. This strategic population sizing mechanism helps the MA algorithm to make its search more focused and more effective.			

Second, from a computational perspective, we apply the proposed method to solving the challenging (NP-hard) critical node problem (CNP). For this purpose, we integrate a dedicated local improvement procedure (named diversified late acceptance search) and a structured crossover (called double-backbone crossover) within the variable population memetic search framework. We demonstrate the competitiveness of the resulting VPMS algorithm on two sets of 42 synthetic and real-world benchmark instances in the literature compared to the best-performing CNP methods. Specifically, our VPMS algorithm matches the best-known results for 23 instances and notably discovers new record results (improved upper bounds) for 13 instances. It is also the first heuristic algorithm able to steadily reach the optimal solutions for all 9 instances with known optima in only one minute.

Finally, the proposed VPMS method is of generic nature and can help to design effective memetic algorithms for solving various (combinatorial) problems. It can also be used to boost an existing MA by integrating within it the strategic population sizing mechanism introduced in this work. As such, it is expected that the VPMS method contributes favorably to better solve numerous optimization problems.

The rest of this paper is organized as follows. Section \ref{Sec:Related Work} presents a brief literature review of studies on population sizing in evolutionary algorithms. Section \ref{Sec:Variable Population Memetic Algorithm} introduces the proposed VPMS approach. Section \ref{Sec:VPMS Applied to The Critical Node Problem} shows the case study of applying the general VPMS approach to the critical node problem, including detailed computational results and comparisons with state-of-the-art CNP algorithms. Finally, Section \ref{Sec:Conclusions and Future Work} summarizes the work with potential research perspectives.

\section{Related work on population control in evolutionary algorithms}
\label{Sec:Related Work}

%In this section, we provide a brief review on studies of population size in general evolutionary algorithms.

Evolutionary algorithms (EAs) are population-based computation methods. One important issue concerning EAs is population control. Indeed this issue has been investigated in a number of studies in the literature mainly on \textit{continuous} optimization \cite{Guan2017}. These studies can be divided into two categories, namely deterministic methods and adaptive methods.

%. We provide a brief review of the main stuntil 2012. Since then the literature Indeed, a number of studies have

%Memetic algorithms are basically population-based hybrid evolutionary algorithms (EAs) by combining population-based methods with local search methods. Much work has been carried out for the control of population size in EAs. These studies can be roughly divided into two categories, namely deterministic methods and adaptive methods.

\textbf{Deterministic methods} change the population size during the evolution process according to some deterministic rules. For example, Fern{\'a}ndez et al. \cite{Fernandez2003} proposed a method based on the phenomena of plague, in which a fixed number of bad individuals are removed at each generation. Instead of removing individuals at each generation, Brest et al. \cite{Brest2012} presented a method, which starts from a small population size. Then, the population is increased with a specific size determined by a constant value and then reduced by half during the evolution. Besides increasing or decreasing a specific number of individuals after each specific number of generations during evolution, a few methods have been proposed to automatically adjust the size of population based on predefined functions. For example, Koumousis et al. \cite{Koumousis2006} introduced functions with saw-tooth shape for adjusting the population size.

\textbf{Adaptive methods} utilize feedback information from the search to determine the direction and magnitude of change of population size. For example, Arabas et al. \cite{Arabas1994} presented a genetic algorithm with variable population, which eliminates the population size as an explicit parameter by using features such as ``age'' and ``maximal lifetime'' of individuals. Eiben et al. \cite{Eiben2004} introduced a technique that grows the population in case of high fitness improvement or long lasting stagnation, while shrinking the population in case of short period stagnation. Besides the fitness of individuals, information on fitness diversity of the population was also used to control the population size. For example, Tirronen and Neri \cite{Tirronen2009} proposed a method based on fitness diversity measured by the distances between pairs of individuals along with their fitness values to control population size.

Although considerable studies have been performed on population control in EAs, existing studies are not fully suitable for memetic algorithms because of the totally different algorithm dynamics \cite{Karapetyan2011}. Moreover, compared to population control in EAs for continuous problems, very few effort has been made on memetic algorithms (MAs) for combinatorial problems. To our knowledge, the only study on population sizing in discrete MAs was presented in \cite{Karapetyan2011}. In their work, Karapetyan and Gutin presented a memetic algorithm for the multidimensional assignment problem, where the population size is adjustable according to a function of the average running time of the local optimization component.

In addition to the scarceness of investigations on population control in MAs for discrete problems, it is surprising to observe that the most recent studies on population control dated back to 2012. In this work, we fill the gap by proposing variable population memetic search (see Section \ref{Sec:Variable Population Memetic Algorithm}) which enhances the memetic search framework with a strategic population sizing mechanism.

\section{Variable population memetic search}
\label{Sec:Variable Population Memetic Algorithm}

In this section, we present the variable population memetic search (VPMS) framework, which introduces a strategic population sizing mechanism into memetic algorithms.

\subsection{General scheme}
\label{SubSec:General Scheme}

%\textcolor[rgb]{1,0,0}{['population size controlling' is replaced by 'population sizing']}

Like any population-based search algorithm, the performance of a memetic algorithm depends critically on its ability of maintaining a suitable balance of exploration and exploitation of the search space. The proposed variable population memetic search (VPMS) framework aims to encourage such a search balance via a dynamic population sizing mechanism.

From a search perspective, the VPMS approach starts with a small population of two individuals (solutions) to favor exploitation and then strategically adjusts the populations size to influence the population diversity and thus the balance of exploitation and exploration.

From an algorithmic perspective, VPMS mainly consists of five components: population building (Section \ref{SubSec:Population Building}), solution construction (Section \ref{SubSec:Solution Construction}), local improvement (Section \ref{SubSec:Local improvement}), population updating (Section \ref{SubSec:Population Updating}) and population sizing (Section \ref{SubSec:Population Size Controling}). As shown in Algorithm \ref{Alg:Pseudo-Code VPMS Approach}, VPMS starts with an elite population of only two solutions that are obtained by the \textit{PopulationBuilding()} procedure (line 4). From this small elite population, VPMS enters a ``while'' loop (lines 8-26) to perform its evolutionary search until a given stopping condition is satisfied. At each generation, two or more parents are selected to create an offspring solution based on the \textit{SolutionConstruction()} procedure (line 10). Afterwards, the offspring solution is further improved by the \textit{LocalImprovement()} procedure (line 12). The improved offspring solution is then inserted into the population according to the \textit{PopulationUpdating()} procedure (line 22). In addition to these basic components of a general memetic algorithm, the proposed VPMS approach specifically integrates a new component to dynamically control the population size according to the \textit{PopulationSizing()} procedure (line 24). With the help of its strategic population sizing mechanism, the algorithm adapts (i.e., increases or decreases) its population size according to the current search status.

\begin{algorithm}[!ht]
\begin{small}
\caption{Variable population memetic search}
\label{Alg:Pseudo-Code VPMS Approach}
 	\KwIn{Problem instance $I$ with a minimization objective $f$.}
 	\KwOut{The best solution $S^*$ found}
 	\Begin{
 	//build an elite population of two solutions; Sect. \ref{SubSec:Population Building} \\
    $ps \leftarrow 2$;\\
    $\textit{P} = \{S_1 ,S_2\} \leftarrow \textit{PopulationBuilding}(ps)$;\\
    //record the best solution\\
    $S^{*} \leftarrow \arg \min_{i \in [1,2]} f(S_i)$;\\
    $\textit{gens} \leftarrow 0$, $\textit{idle\_gens} \leftarrow 0$;\\
	\While{a stopping condition is not reached}
	{
		//construct an offspring solution; Sect. \ref{SubSec:Solution Construction}\\
        $S' \leftarrow \textit{SolutionConstruction}(P)$;\\
		//improve it by local optimization, Sect. \ref{SubSec:Local improvement}\\
        $S' \leftarrow \textit{LocalImprovement}(S', \textit{MaxIdleIters})$;\\
        //update the best solution\\
		\If{$f(S') < f(S^*)$}{
			$S^{*} \leftarrow S'$;\\
            $\textit{idle\_gens} \leftarrow 0$;\\
		}
        \Else{
            $\textit{idle\_gens} \leftarrow \textit{idle\_gens} + 1$;\\
        }
        //update the population; Sect. \ref{SubSec:Population Updating}\\
        $P \leftarrow \textit{PopulationUpdating}(P,S')$;\\
        //control population size; Sect. \ref{SubSec:Population Size Controling}\\
        $P \leftarrow \textit{PopulationSizing}(P,\textit{idle\_gens})$;\\
        $\textit{gens} \leftarrow \textit{gens} + 1$;
	}
}
\Return the best solution found $S^*$;
\end{small}
\end{algorithm}

\subsection{Population building}
\label{SubSec:Population Building}

VPMS uses a \textbf{population building strategy} to build an initial population. Specifically, an initial solution is first generated by a random or greedy construction method, and then improved by a local improvement procedure. A second high quality solution is generated in the same way. Our population building strategy distinguishes itself from the general strategy by using a small population of only two solutions. This is based on two particular considerations. First, building an initial population of multiple high-quality solutions may be time-consuming. In some settings where a time limit is given, the allowable time can fully be consumed during the population building phase, leaving no time for further search (see \cite{Zhou2019} for an example). Second, at the beginning of the search, since the search space is not examined yet, it is desirable to perform an intensified search to locate as fast as possible some first high-quality local optima. %By using a restricted population of only two individuals, VPMS first focuses on exploitation in a time-effective way.

\subsection{Solution construction}
\label{SubSec:Solution Construction}

Solution construction is an important component of a memetic algorithm and forms one leading force for exploration. It aims to create new solutions (offspring) by blending existing solutions. Crossover is a widely-used method to generate offspring solutions, which is responsible for exploring new search areas of the solution space. Crossover operators usually considers two or more parents to form one or more offspring solutions. Various crossover operators have been developed and studied in the literature for different representations \cite{Pavai2016}, such as single point crossover, uniform crossover, partially matched crossover, and order crossovers. In addition to these (general) operators, it is often advantageous to design dedicated crossovers that enable the offspring solutions to inherit meaningful features (building blocks) of the studied problem from the parent solutions. For example, several specific crossovers based on the idea of ``backbone'' were proposed for problems such as graph coloring \cite{Porumbeletal2010,JinHao2019} and critical nodes identification \cite{Zhou2019}. Finally, other solution construction methods have also studied. For example, Martins et al. \cite{Martins2018} developed a pattern based solution construction method, which tries to construct offspring based on common patterns mined from a set of elite solutions.

\subsection{Local improvement}
\label{SubSec:Local improvement}

Local improvement plays a critical role in a memetic algorithm and ensures essentially the role of intensive exploitation of the search space by focusing on a limited region. The local improvement procedure can benefit from many general local search methods \cite{Hoos2004} such as hill climbing, simulated annealing, tabu search, threshold accepting, and variable neighborhood search. Still, these general methods need to be adapted to the given problem in particular by designing suitable search components (e.g., neighborhoods). For example, Segura et al. \cite{Segura2017} integrated a simple stochastic hill climbing into a memetic algorithm for the frequency assignment problem. Benlic and Hao \cite{Benlic2015} used breakout local search as the local improvement procedure of an effective memetic algorithm for quadratic assignment. Tang et al. \cite{Tang2009} applied within their memetic algorithm an extended neighborhood search procedure for capacitated arc routing. Wu et al. \cite{Wu2019} developed a game-based memetic algorithm for vertex cover of networks, where local improvement is based on an asynchronous updating best response rule of snowdrift game. For the critical node problem of Section \ref{Sec:VPMS Applied to The Critical Node Problem}, our local improvement procedure is based on a diversified late acceptance search algorithm \cite{Namazi2018}.

\subsection{Population updating}
\label{SubSec:Population Updating}

For each offspring solution constructed by the solution construction component (Section \ref{SubSec:Solution Construction}) and further improved by the local improvement component (Section \ref{SubSec:Local improvement}), a decision is made to determine whether and how the offspring solution is inserted into the population according to a population updating strategy. For this purpose, existing population replacement strategies can be used in the VPMS approach. For instance, a popular population updating strategy replaces the worst solution if the offspring has a better quality and is distinct from any solution in the population. This greedy strategy however may lead to a premature loss of population diversity, given that only the fitness of the offspring is considered regardless of its distance to the individuals in the population. To better manage the population diversity, more elaborated updating strategies exist in the literature. For example, S{\"{o}}rensen and Sevaux \cite{Sorensen2006} proposed a population management strategy to maintain a healthy diversity of the population, which simultaneously considers offspring's quality and its distance to the individuals in the population.

\subsection{Population sizing}
\label{SubSec:Population Size Controling}

As its key component, our VPMS approach integrates a \textbf{strategic population sizing} mechanism (see Algorithm \ref{Alg:Main Flow of the Strategic Population sizing mechanism}) to dynamically adjust the population size during the evolutionary search. This mechanism is composed of a population expanding strategy (to add new individuals) and a population rebuilding strategy (to shrink the population to two individuals). In general terms, we expand the population with new elite solutions when a search stagnation is detected. If the population becomes too large but the search still stagnates, we reduce the population to two solutions. A search stagnation occurs when the best recorded solution $S^*$ has not been updated after $\textit{MaxIdleGens}$ consecutive generations. %The pseudo code of our strategic population sizing mechanism is provided in .

\begin{algorithm}[!ht]
\begin{small}
\caption{The pseudo code of the strategic population sizing mechanism.}
\label{Alg:Main Flow of the Strategic Population sizing mechanism}
 	\KwIn{Population $P$ of size $ps$, maximum allowable population size $ps_{max}$, population size increment $ps_{inc}$ and counter of generations without improvement $\textit{idle\_gens}$.}
 	\KwOut{A new population $P$.}
 	\Begin{
    \If{$\textit{idle\_gens} > \textit{MaxIdleGens}$}{
 	    //expand the population by adding new solutions;\\
        \If{$ps < ps_{max}$}{
            $ps \leftarrow ps + ps_{inc}$;\\
            $P \leftarrow \textit{PopulationExpanding}(P,ps)$;\\
        }
        //rebuild population based on the best recorded solution;\\
        \Else{
            $ps \leftarrow 2$;\\
            $P \leftarrow \textit{PopulationRebuilding}(S^*,ps)$;\\
        }
        //update the best solution\\
        $S^* \leftarrow \arg \min_{i \in [1,\ldots,ps]} f(S_i)$;\\
        $\textit{idle\_gens} \leftarrow 0$;
    }
}
\Return A new population $P$.
\end{small}
\end{algorithm}

\subsubsection{Population expanding}
\label{SubSubSec:Population Expanding Strategy}

When the search stagnates, we try to break the stagnation by introducing more diversity into the algorithm. It is a common sense that the greater the population size, the greater the population diversity. Therefore, we increase the diversity by expanding the population upon search stagnation. Specifically, our \textbf{population expanding strategy} adds $ps_{inc}$ new high quality solutions into the population, where each new solution is generated according to the population building strategy of Section \ref{SubSec:Population Building} and added to the population  only if the new solution is not the same as any existing solution in the population. %, it is added into the population. Otherwise, it is discarded. The process continues until $ps_{inc}$ new solutions are added into the current population. %In our VPMS approach, this population expanding procedure will be invoked only when a search stagnation is detected.

\subsubsection{Population rebuilding}
\label{SubSubSec:Population Rebuilding Strategy}

Since large populations usually consume more computational resources, we rebuild the population when the population size exceeds an allowable threshold value $ps_{max}$ while the search is still stagnating. Unlike the population building strategy of Section \ref{SubSec:Population Building}, the \textbf{population rebuilding strategy} shrinks the population to a small population of only two solutions. The new population retains always the best recorded solution $S^*$ and includes another elite solution generated in the same way as the population building strategy of Section \ref{SubSec:Population Building}. %In addition to fulfilling the purpose of reducing the computational burden, the population rebuilding strategy is that

% in a new solution is first constructed by in a greedy way or a random way, then it is improved by a local improvement procedure. %Once this new population is built, VPMS will continue to search.

\section{VPMS applied to the critical node problem}
\label{Sec:VPMS Applied to The Critical Node Problem}

This section presents a practical application of the VPMS approach to solve the critical node problem (CNP) and demonstrates its competitiveness compared to the state of the art.

\subsection{Critical node problem}
\label{SubSec:Critical Node Problem}

Let $G = (V,E)$ be an undirected graph with $|V| = n$ nodes and $|E| = m$ edges, the critical node problem (CNP) involves  identifying a subset of nodes $S \subseteq V$ ($|S| \leqslant k$) such that the removal of the vertices in $S$ leads to a residual graph $G[V \setminus V]$ with the minimum pairwise connectivity. These removed nodes are usually called as \textit{critical nodes}. Once the critical nodes have been removed from $G$, the residual graph $G[V \setminus S]$ can be represented by a set of disjoint connected subgraphs (i.e., components) $\mathcal{H} = \{\mathcal{C}_1 ,\mathcal{C}_2 ,\ldots, \mathcal{C}_T\}$, where a connected component $\mathcal{C}_i$ is a set of nodes such that there exists a path from a node to any other node in this component, and no edge exists between any two connected components.

Since any subset $S \subset V$ of $k$ nodes ($k$ is a positive integer) is a feasible solution for the given graph, the search space $\Omega$ is composed of all possible $k$-node subsets of $V$, i.e., $\Omega = \{S \subset V: |S| = k\}$. Clearly this search space has a size of $\binom{n}{k} = \frac{n!}{k!(n-k)!}$, which increases extremely fast with $n$ and $k$.

Recall that $\sum_{i,j \in V} u_{ij}$ is a measure of the total pairwise connectivity of a graph, where $u_{ij} = 1$ if and only if node $i$ and node $j$ are in the same component, otherwise $u_{ij} = 0$. Therefore, the objective function can be rewritten as
\begin{equation}\label{Equ:Objective Function}
    f(S) = \sum_{\mathcal{C}_i \in \mathcal{H}} \frac{|\mathcal{C}_i|(|\mathcal{C}_i|-1)}{2}
\end{equation}
where $S$ is a set of critical nodes, $|\mathcal{C}_i|$ is the size of the $i$-th component of residual graph $G[V \setminus S]$. It is known that $f(S)$ can be easily computed by fast algorithms like breadth or depth first search algorithms in $O(|V|+|E|)$ time using an adjacency list representation of the graph \cite{Cormen2009}.

\begin{figure}[!ht]
\centering
\includegraphics[width=3.5in]{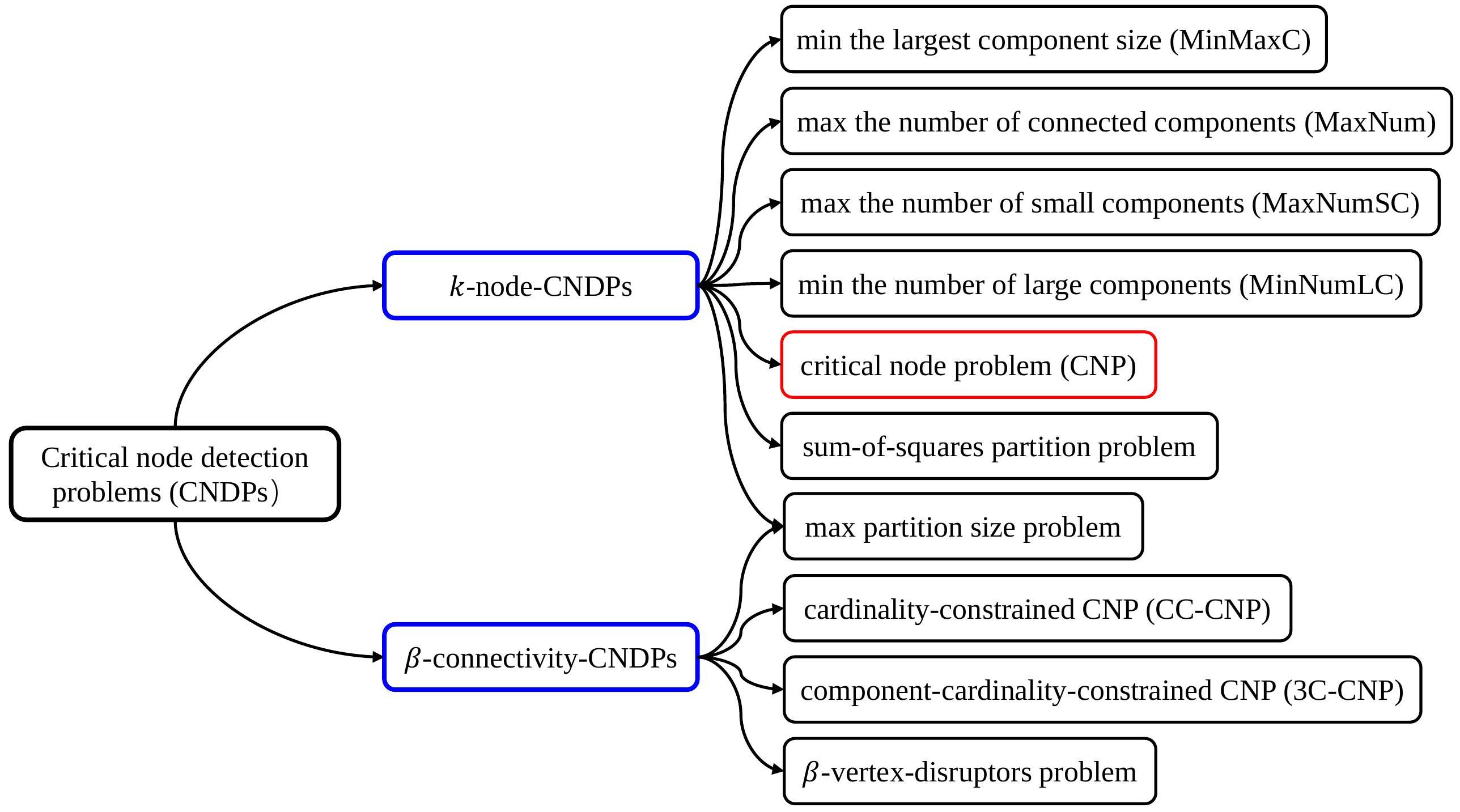}
\caption{A taxonomy of critical node detection problems.}
\label{Fig:CNDP Classification}
\end{figure}

CNP has several interesting variants, which optimize different objectives, such as minimizing the size of the largest connected component and maximizing the number of connected components. A detailed classification of the main CNP variants is provided in Fig.~\ref{Fig:CNDP Classification}, while more details about these variants can be found in the recent survey \cite{Lalou2018}.

\subsection{Existing studies on CNP}
\label{SubSec:Existing Studies on CNP}

Given its practical and theoretical significance, CNP has been widely studied in the literature. Various solution approaches have been developed, which can be divided into two categories: exact algorithms and heuristic algorithms.

\textbf{Exact algorithms} aim to provide the proven optimal solutions. Since they have exponential complexities, they are particularly useful for handling special graphs. For example, Di Summa et al. \cite{Summa2011} proved that CNP is polynomially solvable on trees via dynamic programming. Addis et al. \cite{Addis2013} defined another dynamic programming procedure that solves CNP in polynomial time when the graph has bounded treewidth, which generalizes and extends the results presented in \cite{Summa2011} for the case of a tree. Di Summa et al. \cite{Summa2012} also studied branch and cut algorithms for detecting critical nodes in general graphs, where an integer linear programming model with a non-polynomial number of constraints is proposed. However, the above-mentioned exact algorithms are able to solve CNP on general graphs with no more than 150 nodes. Veremyev et al. \cite{Veremyev2014} developed a more compact linear 0-1 formulations of CNP that requires $n^2$ constraints, which was tested on much larger real-world sparse networks. % with 1500 nodes within 50,000 seconds. It is worth noting that these real-world networks are not real-world benchmark instances used in our following experiments.

\textbf{Heuristic algorithms} aim to find high-quality solutions within reasonable time without guaranteeing the optimality of the solutions found. Heuristic algorithms are particular useful to handle problem instances that cannot be solved by exact algorithms. Heuristic algorithms for CNP can be divided into three categories: constructive approaches, local search approaches and population-based approaches.

\begin{itemize}
\item \textbf{Constructive approaches} start from an ``empty solution'' and repeatedly extend the current solution until a complete solution is obtained. A early constructive heuristic obtains an initial solution by determining a vertex cover, and then uses a greedy rule to add some nodes back (\textit{Add} for short) to the original graph until a feasible solution is obtained \cite{Arulselvan2009}. Inversely, a greedy heuristic starts from an empty set and then removes nodes (\textit{Remove} for short) from the original graph using a greedy rule \cite{Ventresca2014}. Addis et al. \cite{Addis2016} developed several hybrid constructive approaches by alternating the application of these two basic greedy operations. Moreover, a sophisticated multi-start greedy algorithm (CNA1 for short) was developed in \cite{Pullan2015}.

\item \textbf{Local search approaches} start from a complete candidate solution and then try to improve the current solution by performing local moves. For example, Aringhieri et al. \cite{Aringhieri2016b} presented two local search algorithms based on the iterated local search and variable neighborhood search frameworks, respectively. Recently, Zhou and Hao \cite{Zhou2017b} proposed a fast and effective iterated local search (FastCNP for short), which employs an effective two-phase node exchange strategy to locate high-quality solutions and applies a destructive-constructive perturbation procedure to drive the search to new regions when the search stagnates.

\item \textbf{Population-based hybrid approaches} work with multiple solutions that are manipulated by search operators such as recombination and mutation. For example, Ventresca \cite{Ventresca2012} proposed a population-based incremental learning approach for CNP, where a combinatorial unranking-based problem representation is used. Aringhieri et al. \cite{Aringhieri2016a} introduced an efficient evolutionary framework for solving different variants of CNP, where greedy rules are used to guide the search towards good quality solutions during reproduction and mutation phases. Recently, Zhou et al. \cite{Zhou2019} developed an effective memetic search approach (MACNP for short) for both CNP and CC-CNP, which achieves the state-of-the-art performance on benchmark instances from two popular synthetic and real-world datasets (which are presented in Section \ref{SubSubSec:Benchmark Instances}).
\end{itemize}

It is worth noting that the state-of-the-art results on CNP benchmark instances were reported by CAN1 \cite{Pullan2015}, FastCNP \cite{Zhou2017b} and MACNP \cite{Zhou2019}. These algorithms will thus be used as reference algorithms in our comparisons in Section \ref{SubSubSec:Comparisons With State-of-the-Art Algorithms}.

\subsection{Variable population memetic algorithm for CNP}
\label{SubSec:VPMS for CNP}

Our variable population memetic search algorithm for CNP (denoted by VPMS$_{CNP}$) strictly follows Algorithm \ref{Alg:Pseudo-Code VPMS Approach}, while specifying the solution construction component and the local improvement component. Additionally, for its population management, VPMS$_{CNP}$ applies the rank-based quality-and-distance pool updating strategy presented in \cite{Zhou2017a}.

\subsubsection{Double backbone-based crossover}
\label{SubSubSec:Double Backbone-Based Crossover}

Concerning the solution construction component, we adopt the double backbone-based crossover (DBC) \cite{Zhou2019}, which performs structured combinations by inheriting common elements from the parent solutions. Specifically, from two parent solutions $S_1$ and $S_2$ randomly taken from the population, DBC generates an offspring solution in three steps: a) create a partial solution by inheriting the common elements shared by the parents (i.e., identified by the set $S_1 \cap S_2$); b) add the elements from the set $(S_1 \cup S_2) \setminus (S_1 \cap S_2)$ into the partial solution in a probabilistic way; c) repair the solution structurally until a feasible solution is achieved by either adding elements from the set $V \setminus (S_1 \cup S_2)$ or removing elements from the solution. Once a feasible offspring solution is obtained, it is further ameliorated by the diversified late acceptance search procedure below.

\begin{comment}
\begin{algorithm}[!ht]
\begin{small}
\caption{The pseudo code of VPMSCNP$_{DLAS}$ algorithm.}
\label{Alg:Pseudo-Code VPMSCNP}
 	\KwIn{A CNP instance $G$}
 	\KwOut{The best solution $S^*$ found}
 	\Begin{
 	//build an elite population of only two solutions;\\
    $ps \leftarrow 2$;\\
    $P = \{S_1 , S_2\} \leftarrow \textit{PopulationBuilding}(ps)$;\\
    //record the best solution\\
    $S^{*} \leftarrow \arg \min_{i \in [1,2]} f(S_i)$;\\
    $\textit{gens} \leftarrow 0$, $\textit{idle\_gens} \leftarrow 0$;\\
	\While{a stopping condition is not reached}
	{
        //select randomly two parent solutions\\
        $S_i ,S_j \leftarrow \textit{SelectParents}(P)$;\\
		//generate an offspring solution by crossover\\
        $S' \leftarrow \textit{DoubleBackboneBasedCrossover}(S_i ,S_j)$;\\
		//improve it by a diversified late acceptance search\\
        $S' \leftarrow \textit{DiversifiedLateAcceptanceSearch}(S',\textit{MaxIdleIters})$;\\
        //update the best solution\\
		\If{$f(S') < f(S^*)$}{
			$S^{*} \leftarrow S'$;\\
            $\textit{idle\_gens} \leftarrow 0$;\\
		}
        \Else{
            $\textit{idle\_gens} \leftarrow \textit{idle\_gens} + 1$;\\
        }
        //update the population\\
        $P \leftarrow \textit{PopulationUpdating}(P,S')$;\\
        //control population size\\
        $P \leftarrow \textit{PopulationSizeControling}(P,\textit{idle\_gens})$;\\
        $\textit{gens} \leftarrow \textit{gens} + 1$;
	}
}
\Return the best solution found $S^*$;
\end{small}
\end{algorithm}
\end{comment}

\subsubsection{Diversified late acceptance search}
\label{SubSubSec:Diversified Late Acceptance Search}

Diversified late acceptance search (DLAS) \cite{Namazi2018} is an iterative local search algorithm that is inspired by the late acceptance hill climbing (LAHC) algorithm \cite{Burke2017}. Both DLAS and LAHC start their search from an initial solution and iteratively accepts or rejects candidate solutions until a given stopping condition is met. The LAHC method uses a fitness array of size $HL$ (i.e., \textit{history length}) to memorize the cost of the previous encountered solutions. Initially, all elements of this array are filled with the cost of the initial solution $S$. At each subsequent iteration $\textit{iters}$, a candidate solution $S'$ is generated. Then, an acceptance decision is made according to a comparison between the cost of the candidate solution $S'$ and the previous solution cost stored at position $v$ (the virtual beginning of the fitness array, $v \leftarrow \textit{iters} \mod HL$). Specifically, the candidate solution $S'$ is accepted if its cost is not worse than the cost $f_v$ at position $v$ of the fitness array. After the transition from the current solution to the candidate solution (i.e., $S'$ becomes the new current solution), the value of position $v$ of the fitness array, is updated by $f_v \leftarrow f(S')$. The process repeats until the given stopping condition is met.

DLAS (Algorithm \ref{Algorithm:DLAS}) enhances LAHC by increasing the diversity of the accepted solutions and improving the diversity of the values stored in the fitness array. This is achieved by adopting a new acceptance strategy and a new replacement strategy which takes worsening, improving, and sideways movement scenarios into account \cite{Namazi2018} (lines 14-35). Specifically, the new acceptance strategy compares at each iteration the fitness value $f(S')$ of the candidate solution $S'$ with the maximum fitness value $f_{max}$ in the fitness array instead of only comparing it with $f_v$ (lines 14-23). For the new replacement strategy, the replacement occurs only in two cases: 1) if $f(S) > f_v$, and 2) if $f(S) < f_v$ and $f(S)< f_{prev}$ simultaneously hold (line 27). Our experiments showed that the combination of the new acceptance and replacement strategies in DLAS is indeed more effective in increasing the search diversity than just increasing the length of the fitness array, and consequently helps the algorithm to reach high quality solutions in less time.

To generate a candidate solution, DLAS relies on the component-based two-phase node exchange operator (denoted by swap) introduced in \cite{Zhou2019}, which exchanges a node $u \in S$ with a node $v \in V \setminus S$ from a \textit{large component}. Let $G[V \setminus S]$ be the residual graph $G[V \setminus S]$ induced by the current solution $S$ and $\mathcal{H} = \{\mathcal{C}_1 ,\mathcal{C}_2 ,\ldots, \mathcal{C}_T\}$ be the set of connected components of $G[V \setminus S]$. For a  swap operation, we consider as candidate nodes a restricted set of nodes $W \subset V \setminus S$ such that $W = \cup_{|\mathcal{C}_i| \geqslant L}\mathcal{C}_i$, where $L$ is a predefined threshold value to qualify large components in the residual graph. Thus, a candidate neighbor solution $S'$ is obtained by $swap(u ,v)$, where $u \in S$ and $v \in W$. For a given candidate solution, its quality can be evaluated in $O(|V|+|E|)$ time with a modified depth-first search algorithm \cite{Hopcroft1973} according to Eq. (\ref{Equ:Objective Function}).

\begin{algorithm}[!ht]
\begin{small}
\caption{The pseudo code of the DLAS procedure.}
\label{Algorithm:DLAS}
 	\KwIn{Initial solution $S$ and maximum allowable number of idle iterations $\textit{MaxIdleIters}$.}
 	\KwOut{The best solution $S^*$ found}
 	\Begin{
    Initialize the history length $HL$, $S^* \leftarrow S$;\\
    \For{$\forall i \in \{0,\ldots,HL-1\}$}{
        $f_i \leftarrow f(S)$;
    }
    $f_{max} \leftarrow f(S)$, $nbr\_max \leftarrow HL$;\\
    Initialize $\textit{iters} \leftarrow 0$, $\textit{idle\_iters} \leftarrow 0$;\\
	\While{$\textit{idle\_iters} < \textit{MaxIdleIters}$}
	{
        $f_{prev} \leftarrow f(S)$;\\
		$S' \leftarrow \textit{SwapOperation}(S)$;\\
		Calculate its cost function $f(S')$;\\
        /*calculate the virtual beginning*/\\
        $v \leftarrow \textit{iters} \mod HL$;\\
        \If{$f(S') = f(S)$ or $f(S') < f_{max}$}{
            $S \leftarrow S'$, $f(S) \leftarrow f(S')$;\\
            \If{$f(S) < f(S^*)$}{
                $S^* \leftarrow S$, $f(S^*) \leftarrow f(S)$;\\
                $\textit{idle\_iters} \leftarrow 0$;
            }
            \Else{
                $\textit{idle\_iters} \leftarrow \textit{idle\_iters} + 1$;
            }
        }
        \If{$f(S) > f_v$}{
            $f_v \leftarrow f(S)$;
        }
        \ElseIf{$f(S) < f_v$ and $f(S) < f_{prev}$}{
            \If{$f_v = f_{max}$}{
                $\textit{nbr\_max} \leftarrow \textit{nbr\_max} - 1$;
            }
            $f_v \leftarrow f(S)$;\\
            \If{$\textit{nbr\_max} = 0$}{
                compute $f_{max}, \textit{nbr\_max}$;\\
            }
        }
        $\textit{iters} \leftarrow \textit{iters} + 1$;
	}
}
\Return the best solution found $S^*$;
\end{small}
\end{algorithm}

\subsection{Computational studies of VPMS for CNP}
\label{SubSec:Computational results}

This section is devoted to an experimental evaluation of the performance of the VPMS$_{CNP}$ algorithm in comparison with state-of-the-art CNP algorithms.

\subsubsection{Benchmark instances}
\label{SubSubSec:Benchmark Instances}

Our computational experiments were performed on two widely-used benchmark datasets: synthetic dataset\footnote{Available at \url{http://individual.utoronto.ca/mventresca/cnd.html}} and real-world dataset\footnote{Available at \url{http://www.di.unito.it/~aringhie/cnp.html}}. The \textbf{synthetic dataset} presented in \cite{Ventresca2012} is composed of 16 graphs with various structures. The \textbf{real-world dataset} introduced in \cite{Aringhieri2016a} includes 26 instances from different practical applications. More details about these two datasets are provided in Table \ref{Tab:Synthetic and Realworld Dataset}.

\begin{table}[!htbp]
\caption{Characteristics of the synthetic and real-world datasets used in the experiments.}
\label{Tab:Synthetic and Realworld Dataset}
\begin{center}
\begin{scriptsize}
\begin{tabular}{lrrr|lrrr}
\toprule[0.75pt]
Instance &$|V|$ & $|E|$ & $k$ & Instance &$|V|$ & $|E|$ & $k$\\
\midrule[0.5pt]
BA500 &500&499&50   &FF250 &250&514&50    \\
BA1000&1000&999&75  &FF500 &500&828&110   \\
BA2500&2500&2499&100&FF1000&1000&1817&150 \\
BA5000&5000&4999&150&FF2000&2000&3413&200 \\
ER235 &235&350&50   &WS250 &250&1246&70   \\
ER466 &466&700&80   &WS500 &500&1496&125  \\
ER941 &941&1400&140 &WS1000&1000&4996&200 \\
ER2344&2344&3500&200&WS1500&1500&4498&265 \\
\midrule[0.5pt]
Bovine        &121&190&3            &Ham3000c &3000&5996&300    \\
Circuit       &252&399&25           &Ham3000d &3000&5993&300    \\
E.coli        &328&456&15           &Ham3000e &3000&5996&300    \\
USAir97       &332&2126&33          &Ham4000  &4000&7997&400    \\
humanDisea    &516&1188&52          &Ham5000  &5000&9999&500    \\
Treni\_Roma   &255&272&26           &powergrid&4941&6594&494    \\
EU\_flights   &1191&31610&119       &Oclinks  &1899&13838&190   \\
openflights   &1858&13900&186       &facebook &4039&88234&404   \\
yeast1        &2018&2705&202        &grqc     &5242&14484&524   \\
Ham1000       &1000&1998&100        &hepth    &9877&25973&988   \\
Ham2000       &2000&3996&200        &hepph    &12008&118489&1201\\
Ham3000a      &3000&5999&300        &astroph  &18772&198050&1877\\
Ham3000b      &3000&5997&300        &condmat  &23133&93439&2313 \\
\bottomrule[0.75pt]
\end{tabular}
\end{scriptsize}
\end{center}
\end{table}

\subsubsection{Experimental settings}
\label{SubSubSec:Experimental Settings}

All our algorithms\footnote{The best solution certificates and our programs will be made available at
\url{http://www.info.univ-angers.fr/pub/hao/VPMS.html}} were implemented in the C++ programming language, and complied using GNU gcc 4.1.2 with `-O3' option on an Intel E5-2670 with 2.5GHz and 2GB RAM under Linux. With a `-O3' flag, running the DIMACS machine benchmark program dfmax\footnote{Available at dfmax: \url{ftp://dimacs.rutgers.edu/pub/dsj/clique}} on our machine requires 0.19, 1.17 and 4.54 seconds to solve graphs r300.5, r400.5 and r500.5 respectively.

\begin{table}[!ht]
\caption{Parameter settings of the proposed VPMS$_{CNP}$ algorithm.}
\label{Tab:Parameter Settings}
\begin{center}
\begin{scriptsize}
\begin{tabular}{llr}
\toprule[0.75pt]
Parameter      & Description & Value\\
\midrule[0.5pt]
$ps_{max}$     & maximal population size & 20\\
$ps_{inc}$     & incremental population size & 2\\
$\textit{MaxIdleGens}$  & maximum number of idle generations & 100\\
$\textit{MaxIdleIters}$ & maximum number of idle iterations in DLAS & 1000\\
\bottomrule[0.75pt]
\end{tabular}
\end{scriptsize}
\end{center}
\end{table}

In the following experiments, we use the well-known two-tailed sign test to check the statistical significance of our comparisons between two algorithms on each comparison indicator. This statistical test is based on the number of instances on which an algorithm is the overall winner, and it is highly recommended in \cite{Demvsar2006}. There are $N = 42$ benchmark instances in our experiments. At a significant level of 0.05, the critical value is $CV^{42}_{0.05} = N/2+1.96\sqrt{N}/2 \approx 27$. This means that algorithm $A$ significantly outperforms algorithm $B$ if $A$ wins at least 27 out of 42 instances.

\subsubsection{Effectiveness of the strategic population sizing mechanism}
\label{SubSubSec:Effectiveness of Strategic Population Size}

Compared to the conventional memetic algorithm framework, the proposed VPMS$_{CNP}$ algorithm integrates a strategic population sizing mechanism to dynamically adjust the population size during the evolutionary search. To verify the effectiveness of our population sizing mechanism, we compare VPMS$_{CNP}$ with an alternative algorithm named FPMS$_{CNP}$ whose population size is fixed to the maximal population size of VPMS$_{CNP}$ while keeping the other components as the same as VPMS$_{CNP}$. As such, FPMS$_{CNP}$ is a classical memetic algorithm which is quite similar to the powerful state-of-the-art memetic algorithm MACNP of \cite{Zhou2019} where a different local improvement procedure is used.

To make a fair comparison between VPMS$_{CNP}$ and FPMS$_{CNP}$, we ran them on the same computing platform with the setting shown in Table \ref{Tab:Parameter Settings}. We independently solved each instance 30 times with different random seeds, and the time limit of each run was limited to $t_{max} =3600$ seconds. Detailed comparative results for both synthetic and real-world datasets are summarized in Tables \ref{Tab:Comparative Performance Between MACNP and VPMSCNP Under Time Limit 3600 Seconds}.

\begin{table*}[!hbtp]
\caption{Comparison of VPMS$_{CNP}$ (with a variable population) against FPMS$_{CNP}$ (with a fixed population) under $t_{max} = 3600$ seconds.}
\label{Tab:Comparative Performance Between MACNP and VPMSCNP Under Time Limit 3600 Seconds}
\begin{center}
\begin{scriptsize}
\begin{threeparttable}
\begin{tabular}{lrrrrrrcrrrrr}
\toprule[0.75pt]
\multicolumn{2}{c}{} & \multicolumn{5}{c}{FPMS$_{CNP}$} && \multicolumn{5}{c}{\textbf{VPMS$_{CNP}$}}\\
\cmidrule[0.5pt]{3-7} \cmidrule[0.5pt]{9-13}
Instance & $f_{bkv}$ &$f_{best}$&$f_{avg}$&$t_{avg}$& $\#gens$ & $\#succ$ &&$f_{best}$&$f_{avg}$&$t_{avg}$&$\#gens$ & $\#succ$\\
\midrule[0.5pt]
BA500 &195\tnote{$\ast$}&\textbf{195}&\textbf{195.0} &0.0 &0 &\textbf{30}&&\textbf{195}&\textbf{195.0}&0.0&0&\textbf{30}\\
BA1000&558\tnote{$\ast$}&\textbf{558}& 558.1 &0.0 &0 &29&&\textbf{558}&\textbf{558.0}&2.4&27&\textbf{30}\\
BA2500&3704\tnote{$\ast$}&\textbf{3704}&3704.6 &2.8 &6 &29&&\textbf{3704}&\textbf{3704.0}&7.2&117&\textbf{30}\\
BA5000&10196\tnote{$\ast$}&\textbf{10196}&\textbf{10196.0} &21.3 &6 &\textbf{30}&&\textbf{10196}&\textbf{10196.0}&10.4&50&\textbf{30}\\
ER235 &295\tnote{$\ast$}&\textbf{295}&\textbf{295.0} &13.6 &3539 &\textbf{30}&&\textbf{295}&\textbf{295.0}&2.0&435&\textbf{30}\\
ER466 &1524&\textbf{1524}&\textbf{1524.0} &45.0 &5181 &\textbf{30}&&\textbf{1524}&\textbf{1524.0}&30.3&3111&\textbf{30}\\
ER941 &5012&\textbf{5012}&5034.0 &442.5 &25209 &\textbf{5}&&\textbf{5012}&\textbf{5026.5}&459.2&22890&3\\
ER2344&902498&912875&\textbf{931976.9} &2456.7 &18838 &1&&\textbf{904113}&933943.7&3012.8&15202&1\\
FF250 &194\tnote{$\ast$}&\textbf{194}&\textbf{194.0} &8.9 &23610 &\textbf{30}&&\textbf{194}&\textbf{194.0}&0.0&0&\textbf{30}\\
FF500 &257\tnote{$\ast$}&\textbf{257}&257.3 &5.0 &4299 &28&&\textbf{257}&\textbf{257.0}&0.5&50&\textbf{30}\\
FF1000&1260\tnote{$\ast$}&\textbf{1260}&1262.3 &354.1 &17751 &16&&\textbf{1260}&\textbf{1260.0}&11.7&554&\textbf{30}\\
FF2000&4545\tnote{$\ast$}&\textbf{4545}&4547.8 &20.5 &402 &13&&\textbf{4545}&\textbf{4545.0}&43.9&1851&\textbf{30}\\
WS250 &3083&\textbf{3083}&3093.3 &1397.5 &63236 &23&&\textbf{3083}&\textbf{3083.1}&1081.5&52449&\textbf{29}\\
WS500 &2072&2078&2089.5 &249.3 &21014 &1&&\textbf{2072}&\textbf{2083.1}&366.7&25120&4\\
WS1000&109807&\textbf{109677}\tnote{$\star$}&\textbf{126764.6} &2629.1 &17445 &1&&119444&134475.5&1506.9&6696&1\\
WS1500&13098&13146&13329.1 &1873.6 &85821 &1&&\textbf{13098}&\textbf{13161.5}&2114.9&31819&9\\
\midrule[0.5pt]
Bovine&268&\textbf{268}&\textbf{268.0}&0.0&0&\textbf{30}&&\textbf{268}&\textbf{268.0}&0.0&0&\textbf{30}\\
Circuit&2099&\textbf{2099}&\textbf{2099.0}&1.3&313&\textbf{30}&&\textbf{2099}&\textbf{2099.0}&1.0&229&\textbf{30}\\
Ecoli&806&\textbf{806}&\textbf{806.0}&0.0&0&\textbf{30}&&\textbf{806}&\textbf{806.0}&0.0&8&\textbf{30}\\
USAir97&4336&\textbf{4336}&\textbf{4897.2}&1126.8&60012&\textbf{12}&&\textbf{4336}&5075.6&1159.5&37122&7\\
humanDisea&1115&\textbf{1115}&1115.3&3.1&292&29&&\textbf{1115}&\textbf{1115.0}&1.6&180&\textbf{30}\\
Treni\_Roma&918&\textbf{918}&\textbf{918.0}&29.7&10216&\textbf{30}&&\textbf{918}&\textbf{918.0}&1.8&765&\textbf{30}\\
EU\_flights&348268&\textbf{348268}&351323.0&74.3&77&2&&\textbf{348268}&\textbf{349265.6}&1145.4&2319&\textbf{18}\\
openflights&26842&26842&28845.3&1812.7&7313&1&&\textbf{26785}\tnote{$\star$}&\textbf{27327.0}&2391.7&9806&2\\
yeast1&1412&\textbf{1412}&\textbf{1412.0}&18.1&104&\textbf{30}&&\textbf{1412}&\textbf{1412.0}&35.9&437&\textbf{30}\\
Ham1000&306349&308731&311422.8&2431.2&22374&1&&\textbf{307117}&\textbf{311169.4}&2027.2&13862&1\\
Ham2000&1243859&\textbf{1244335}&1257388.5&2545.1&9134&1&&1247652&\textbf{1256573.8}&3109.8&7229&1\\
Ham3000a&2844393&2841106\tnote{$\star$}&2861888.3&2553.6&4859&1&&\textbf{2840941}\tnote{$\star$}&\textbf{2859284.4}&3084.4&4660&1\\
Ham3000b&2841270&\textbf{2839733}\tnote{$\star$}&2860997.6&2542.4&4964&1&&2839893\tnote{$\star$}&\textbf{2860810.9}&3179.3&4538&1\\
Ham3000c&2838429&2836076\tnote{$\star$}&2848545.9&2313.0&4411&1&&\textbf{2832073}\tnote{$\star$}&\textbf{2844324.3}&2819.7&4080&1\\
Ham3000d&2831311&\textbf{2830098}\tnote{$\star$}&\textbf{2854757.2}&2903.8&5093&1&&2830291\tnote{$\star$}&2857201.4&3090.1&4608&1\\
Ham3000e&2847909&\textbf{2846371}\tnote{$\star$}&\textbf{2866095.2}&2106.1&3943&1&&2846731\tnote{$\star$}&2867000.6&3231.6&4816&1\\
Ham4000&5044357&\textbf{5060754}&5143157.3&2813.4&3132&1&&5082521&\textbf{5141804.3}&3404.7&3705&1\\
Ham5000&7972525&\textbf{7986458}&\textbf{8098821.1}&3034.5&1943&1&&8011565&8151850.1&3214.4&2970&1\\
powergrid&15862&15899&15954.5&1343.6&10222&1&&\textbf{15873}&\textbf{15909.2}&2964.3&15205&1\\
Oclinks&611326&614467&615030.0&601.6&1276&2&&\textbf{611254}\tnote{$\star$}&\textbf{614296.3}&1658.4&4229&1\\
facebook&420334&703330&798567.9&2708.3&5219&1&&\textbf{691232}&\textbf{780429.1}&3397.0&3753&1\\
grqc&13596&13612&13647.2&802.3&2957&1&&\textbf{13603}&\textbf{13615.5}&2499.9&6367&2\\
hepth&106397&\textbf{107440}&\textbf{109304.9}&2700.6&2459&1&&107939&110158.4&3206.6&2198&1\\
hepph&6156536&9327422&10712034.3&3491.3&7&1&&\textbf{7883063}&\textbf{8689170.1}&3423.8&565&1\\
astroph&53963375&61928888&63311361.7&1684.9&0&1&&\textbf{58322396}&\textbf{59563941.1}&2721.5&225&1\\
condmat&2298596&10352129&10823216.8&1682.5&0&1&&\textbf{6843993}&\textbf{7813436.7}&3388.5&414&1\\
%\midrule[0.5pt]
%better/equal/worse & & 8/20/14 & 7/10/25 & & & 2/10/8 && 14/20/8 & \textbf{25/10/7} & & & 8/10/2 \\
\bottomrule[0.75pt]
\end{tabular}
\begin{tablenotes}
    \item[$\ast$] Optimal solutions obtained by a branch-and-cut algorithm \cite{Summa2012} within 5 days.
    \item [$\star$] Improved best upper bounds.
\end{tablenotes}
\end{threeparttable}
\end{scriptsize}
\end{center}
\end{table*}

In Table \ref{Tab:Comparative Performance Between MACNP and VPMSCNP Under Time Limit 3600 Seconds}, columns 1 and 2 present for each instance its name (Instance) and the best-known value ($f_{bkv}$) reported in the literature \cite{Pullan2015,Aringhieri2016b,Zhou2019}. Columns 3-7 report the results of the FPMS$_{CNP}$ algorithm, namely the best objective value ($f_{best}$) found during 30 runs, the average objective value ($f_{avg}$), the average running time per run to attain a best objective value ($t_{avg}$), the average number of generations per run required to find the best objective value ($\#gens$), and the number of times to successfully find the best objective value ($\#succ$). Similarly, columns 8-12 give the results of VPMS$_{CNP}$. The best values of the compared results in terms of $f_{best}$ and $f_{avg}$ are indicated in bold. For the $\#succ$ indicator, we compare them only when the same $f_{best}$ values are obtained by the two algorithms. %In addition, we also indicate in the last row the number of instances on which an algorithm achieves a better/equal/worse result compared to the other algorithm.

From Table \ref{Tab:Comparative Performance Between MACNP and VPMSCNP Under Time Limit 3600 Seconds}, we observe that the VPMS$_{CNP}$  algorithm (with a variable population) achieves better results on 14 instances, equal results on 20 instances and worse results on 8 instances in terms of $f_{best}$ compared to the fixed population algorithm FPMS$_{CNP}$. However, there is no significant difference between these two algorithms (i.e., $24 < CV^{42}_{0.05}$). For the $f_{avg}$ indicator, VPMS$_{CNP}$ attains better results on 25 instances, equal results on 10 instances and worse results on 7 instances. At a significant level of 0.05, we find that VPMS$_{CNP}$  is significantly better than FPMS$_{CNP}$ on the $f_{avg}$ indicator (i.e., $30 > CV^{42}_{0.05} = 27$). Although VPMS$_{CNP}$ and FPMS$_{CNP}$ achieve the same $f_{best}$ values for 20 out of 42 synthetic instances, VPMS$_{CNP}$ attains these results with a higher success rate on 8 instances, an equal success rate on 10 instances, a lower success rate only on two instance. It is worth noting that VPMS$_{CNP}$ is the first heuristic to steadily ($100\%$) reach the optimal solutions for all 9 instances with known optima (marked by ``$\ast$'' in Table \ref{Tab:Comparative Performance Between MACNP and VPMSCNP Under Time Limit 3600 Seconds}) in only one minute. For the last three large instances, with the help of a variable population, our VPMS$_{CNP}$ algorithm is able to attain better results. Finally, compared to the $f_{bkv}$ values of all 42 benchmark instances, these two algorithms together improve on the best-known results (new upper bounds) on 8 instances (marked by ``$\star$'') and match the best-known upper bounds on 22 instances. These results disclose thus the first positive indications of our strategic population sizing mechanism.

To further study the behavior of the proposed VPMS$_{CNP}$ algorithm, we also report comparative results between VPMS$_{CNP}$ and FPMS$_{CNP}$ with a longer time limit $t_{max} = 7200$ seconds. The detailed computational results are summarized in Table \ref{Tab:Comparative Performance Between MACNP and VPMSCNP Under Time Limit 7200 Seconds}. Table \ref{Tab:Comparative Performance Between MACNP and VPMSCNP Under Time Limit 7200 Seconds} shows that VPMS$_{CNP}$  and FPMS$_{CNP}$  are able to reach better performances. Importantly, the performance difference between VPMS$_{CNP}$  and FPMS$_{CNP}$ is more obvious than the results shown in Table \ref{Tab:Comparative Performance Between MACNP and VPMSCNP Under Time Limit 3600 Seconds}. Specifically, we find that VPMS$_{CNP}$ significantly outperforms FPMS$_{CNP}$ in terms of both $f_{best}$ (i.e., $27 \geqslant CV^{42}_{0.05}$) and $f_{avg}$ indicators (i.e., $31.5 > CV^{42}_{0.05}$). We also observe that these algorithms are able to find new upper bounds on 12 instances (marked by ``$\star$'') and match the best-known upper bounds on 23 instances. These findings indicate that thank to the use of a strategically adjusted population size, the VPMS$_{CNP}$ algorithm is able to use its given computational budget more efficiently and more effectively to find high-quality solutions. This experiment (with both cutoff time limits) also indicates the that the DLAS procedure is an effective local improvement procedure.

\begin{table*}[!hbtp]
\caption{Comparison of VPMS$_{CNP}$ against FPMS$_{CNP}$ under $t_{max} = 7200$ seconds.}
\label{Tab:Comparative Performance Between MACNP and VPMSCNP Under Time Limit 7200 Seconds}
\begin{center}
\begin{scriptsize}
\begin{threeparttable}
\begin{tabular}{lrrrrrrcrrrrr}
\toprule[0.75pt]
\multicolumn{2}{c}{} & \multicolumn{5}{c}{FPMS$_{CNP}$} && \multicolumn{5}{c}{\textbf{VPMS$_{CNP}$}}\\
\cmidrule[0.5pt]{3-7} \cmidrule[0.5pt]{9-13}
Instance & $f_{bkv}$ &$f_{best}$&$f_{avg}$&$t_{avg}$& $\#gens$ & $\#succ$ &&$f_{best}$&$f_{avg}$&$t_{avg}$&$\#gens$ & $\#succ$\\
\midrule[0.5pt]
BA500&195&\textbf{195}&\textbf{195.0}&0.0&0&\textbf{30}&&\textbf{195}&\textbf{195.0}&0.0&0&\textbf{30}\\
BA1000&558&\textbf{558}&558.1&0.2&39&29&&\textbf{558}&\textbf{558.0}&0.3&4&\textbf{30}\\
BA2500&3704&\textbf{3704}&\textbf{3704.0}&3.3&11&\textbf{30}&&\textbf{3704}&\textbf{3704.0}&6.7&47&\textbf{30}\\
BA5000&10196&\textbf{10196}&\textbf{10196.0}&31.4&7&\textbf{30}&&\textbf{10196}&\textbf{10196.0}&11.8&58&\textbf{30}\\
ER235&295&\textbf{295}&\textbf{295.0}&97.1&20535&\textbf{30}&&\textbf{295}&\textbf{295.0}&2.2&536&\textbf{30}\\
ER466&1524&\textbf{1524}&\textbf{1524.0}&42.1&5178&\textbf{30}&&\textbf{1524}&\textbf{1524.0}&28.1&3180&\textbf{30}\\
ER941&5012&\textbf{5012}&5029.2&254.4&16860&\textbf{5}&&\textbf{5012}&\textbf{5017.0}&1754.7&91144&4\\
ER2344&902498&\textbf{902875}&\textbf{927689.7}&4815.0&35997&1&&906904&927865.4&5221.2&27003&1\\
FF250&194&\textbf{194}&\textbf{194.0}&0.0&0&\textbf{30}&&\textbf{194}&\textbf{194.0}&0.0&0&\textbf{30}\\
FF500&257&\textbf{257}&257.3&243.5&29799&26&&\textbf{257}&\textbf{257.0}&0.6&40&\textbf{30}\\
FF1000&1260&\textbf{1260}&1260.2&500.8&30782&24&&\textbf{1260}&\textbf{1260}&12.2&545&\textbf{30}\\
FF2000&4545&\textbf{4545}&4546.5&862.9&51211&8&&\textbf{4545}&\textbf{4545.0}&66.3&1549&\textbf{30}\\
WS250&3083&\textbf{3083}&\textbf{3083.1}&1483.0&70537&28&&\textbf{3083}&3085.1&1199.2&42897&\textbf{29}\\
WS500&2072&\textbf{2072}&2088.3&338.7&56525&2&&\textbf{2072}&\textbf{2083.1}&403.8&20944&\textbf{4}\\
WS1000&109807&\textbf{109712}\tnote{$\star$}&\textbf{126642.7}&4859.6&36025&1&&119795&131959.1&3701.7&17107&1\\
WS1500&13098&13103&13287.9&2916.2&150386&1&&\textbf{13098}&\textbf{13153.2}&3915.5&48240&11\\
\midrule[0.5pt]
Bovine&268&\textbf{268}&\textbf{268.0} &0.0 &0&\textbf{30}&&\textbf{268}&\textbf{268.0} &0.0 &0&\textbf{30}\\
Circuit&2099&\textbf{2099}&\textbf{2099.0} &7.0 &1927&\textbf{30}&&\textbf{2099}&\textbf{2099.0} &1.1 &289&\textbf{30}\\
Ecoli.txt&806&\textbf{806}&\textbf{806.0} &0.0 &0&\textbf{30}&&\textbf{806}&\textbf{806.0} &0.0 &4&\textbf{30}\\
USAir97&4336&\textbf{4336}&\textbf{4665.0} &2886.5 &105951&\textbf{20}&&\textbf{4336}&5060.3 &3626.5 &109348&6\\
humanDisea&1115&\textbf{1115}&\textbf{1115.0} &2.9 &298&\textbf{30}&&\textbf{1115}&\textbf{1115.0} &0.5 &50&\textbf{30}\\
Treni\_Roma&918&\textbf{918}&\textbf{918.0} &7.8 &21591&\textbf{30}&&\textbf{918}&\textbf{918.0} &0.6 &222&\textbf{30}\\
EU\_flights&348268&348269&351657.1 &295.7 &771&1&&\textbf{348268}&\textbf{348434.3} &2307.4 &5682&28\\
openflights&26842&26842&28688.7 &3284.2 &12580&1&&\textbf{26783}\tnote{$\star$}&\textbf{26919.0} &4186.2 &17090&1\\
yeast1&1412&\textbf{1412}&1412.4 &16.9 &60&26&&\textbf{1412}&\textbf{1412.0} &26.5 &324&\textbf{30}\\
Ham1000&306349&308198&310580.2 &4275.0 &39910&1&&\textbf{306349}&\textbf{309912.0} &4055.1 &30476&3\\
Ham2000&1243859&1243289\tnote{$\star$}&1256645.8 &4692.5 &20794&1&&\textbf{1242792}\tnote{$\star$}&\textbf{1251189.7} &5223.8 &14935&1\\
Ham3000a&2844393&2842100\tnote{$\star$}&2855766.8 &4163.2 &9260&1&&\textbf{2840690}\tnote{$\star$}&\textbf{2847291.7} &4776.0 &7607&1\\
Ham3000b&2841270&2838531\tnote{$\star$}&2845347.5 &4466.6 &9347&1&&\textbf{2837584}\tnote{$\star$}&\textbf{2843768.2} &4122.4 &6310&1\\
Ham3000c&2838429&2836053\tnote{$\star$}&2846084.9 &3523.5 &8815&1&&\textbf{2835860}\tnote{$\star$}&\textbf{2839192.3} &4203.1 &6978&1\\
Ham3000d&2831311&\textbf{2827366}\tnote{$\star$}&2847582.4 &4413.3 &10764&1&&2829102\tnote{$\star$}&\textbf{2841551.0} &5631.8 &8778&1\\
Ham3000e&2847909&2844721\tnote{$\star$}&2856464.3 &4286.6 &10228&1&&\textbf{2843000}\tnote{$\star$}&\textbf{2847442.4} &4263.2 &6881&1\\
Ham4000&5044357&5051404&5120450.3 &4405.7 &6449&1&&\textbf{5038611}\tnote{$\star$}&\textbf{5091745.6} &6416.5 &6690&1\\
Ham5000&7972525&\textbf{7968669}\tnote{$\star$}&8078656.1 &4840.9 &4582&1&&7969845\tnote{$\star$}&\textbf{8042058.9} &6276.3 &5383&1\\
powergrid&15862&15908&15957.8 &2148.8 &16566&1&&\textbf{15868}&\textbf{15886.1} &5594.1 &28573&1\\
Oclinks&611326&613430&615029.9 &896.0 &2097&1&&\textbf{611260}\tnote{$\star$}&\textbf{614220.9} &2992.5 &7709&1\\
facebook&420334&676712&793272.9 &5097.0 &13767&1&&\textbf{669910}&\textbf{738856.5} &6537.4 &8250&1\\
grqc&13596&13607&13642.5 &1221.9 &6061&1&&\textbf{13592}\tnote{$\star$}&\textbf{13602.4} &4743.4 &14404&1\\
hepth&106397&106814&109092.4 &3665.1 &4509&1&&\textbf{106792}&\textbf{108673.4} &6378.2 &4633&1\\
hepph&6156536&\textbf{6709598}&\textbf{7541345.1} &6999.4 &182&1&&7211646&7960148.5 &6710.5 &1709&1\\
condmat&53963375&7810704&9508083.3 &6027.8 &7&1&&\textbf{56229708}&\textbf{57421239.1} &6364.6 &592&1\\
astroph&2298596&62281904&63073287.1 &4383.1 &0&1&&\textbf{6057949}&\textbf{6593803.2} &6702.4 &1199&1\\
%\midrule[0.5pt]
%better/equal/worse & & 5/20/17& 5/11/26 & & & 2/10/7 && \textbf{17/20/5} & \textbf{26/11/5} & & & \textbf{7/10/2} \\
\bottomrule[0.75pt]
\end{tabular}
\begin{tablenotes}
    \item [$\star$] Improved best upper bounds.
\end{tablenotes}
\end{threeparttable}
\end{scriptsize}
\end{center}
\end{table*}

\subsubsection{Using the strategic population sizing mechanism to enhance a memetic algorithm}
\label{SubSubSec:Benefit of Strategic Population Size for MACNP}

\begin{figure}[!ht]
\centering
\includegraphics[width=3.5in]{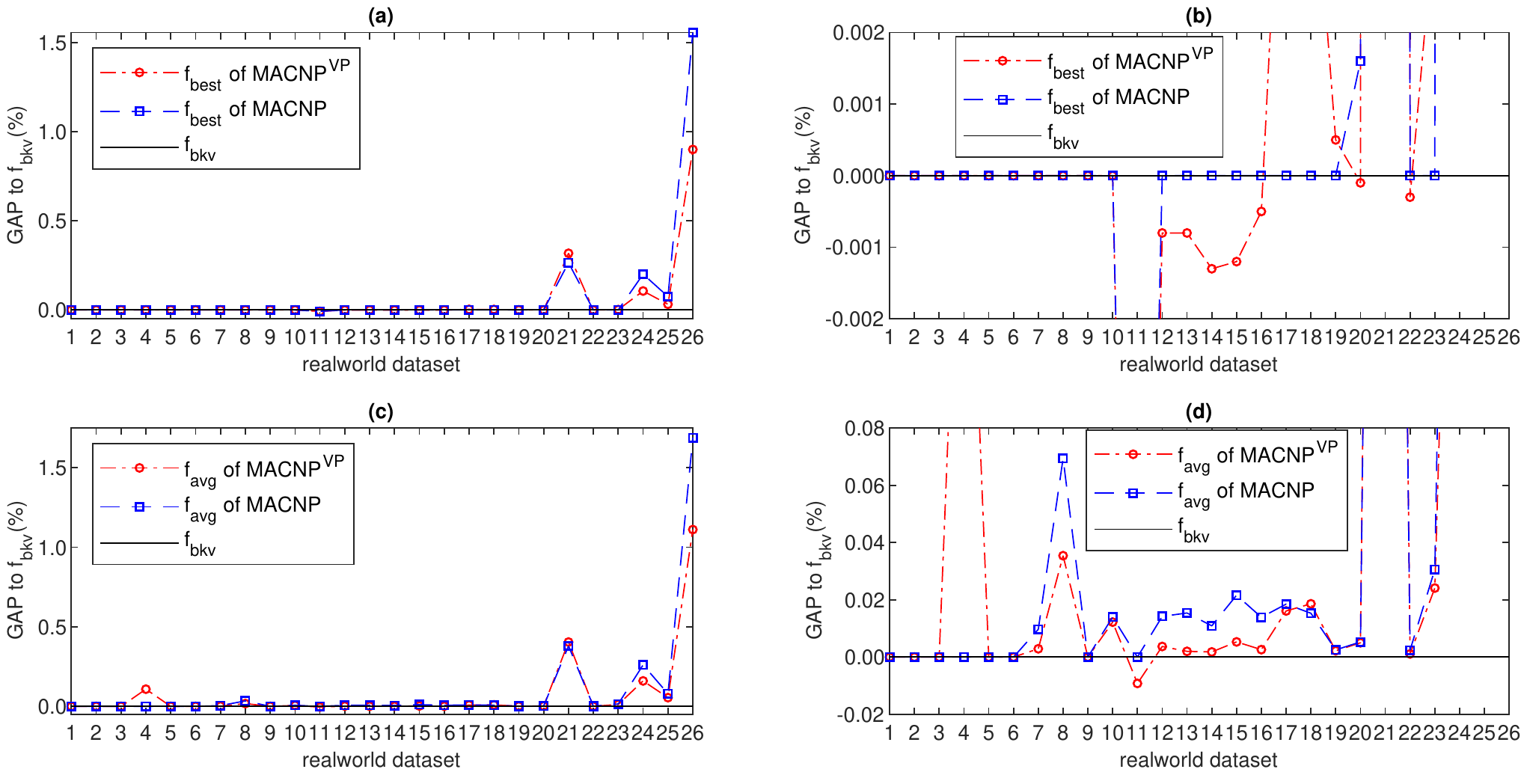}
\caption{\textbf{Comparison between MACNP and MACNP$^{VP}$ under the time limit $t_{max} = 3600$ seconds.} Sub-figures (a) and (b) present the same results under different ranges of $y$-axis. Sub-figures (c) and (d) present the same results under different ranges of $y$-axis.}
\label{Fig:Comparison Between MACNP and VPMSCNP}
\end{figure}

MACNP \cite{Zhou2019} is a recent state-of-the-art memetic search approach for both CNP and CC-CNP. We verify now whether the strategic population sizing mechanism can enhance the performance of this memetic algorithm. For this purpose, we replace the fixed population of MACNP by the strategic population sizing mechanism and we use MACNP$^{VP}$ to denote the resulting MACNP variant with a variable population. We experimentally compare the original MACNP algorithm (with fixed population) and MACNP$^{VP}$ (with a variable population), based on the 26 real-world benchmark instances. We run both algorithm 30 times on each instance under the time limit $t_{max} = 3600$ seconds. The comparative results in terms of the $f_{best}$ and $f_{avg}$ indicators are shown in Fig. \ref{Fig:Comparison Between MACNP and VPMSCNP}. The $x$-axis indicates the instances (named by integer numbers), and the $y$-axis presents the gap of $f$ ($f_{best}$ or $f_{avg}$) values to the best-known values $f_{bkv}$, i.e., $(f-f_{bkv})/f_{bkv}$. Therefore, a negative gap value indicates an improved best upper bound.

From Fig. \ref{Fig:Comparison Between MACNP and VPMSCNP}, we observe that the variable population algorithm  MACNP$^{VP}$  significantly outperforms the fixed population algorithm MACNP in terms of both $f_{best}$ and $f_{avg}$. Specifically, Fig. \ref{Fig:Comparison Between MACNP and VPMSCNP}(a) indicates that MACNP$^{VP}$ achieves better $f_{best}$ values than MACNP except for the $21$-th instance (i.e., \textit{facebook}). A close look of these results (as shown in  \ref{Fig:Comparison Between MACNP and VPMSCNP}(b)) shows that MACNP$^{VP}$ achieves eight new upper bounds. Additionally, MACNP$^{VP}$ also achieves better $f_{avg}$ values than MACNP except for the fourth instance (i.e., \textit{USAir}) (see Fig. \ref{Fig:Comparison Between MACNP and VPMSCNP}(c)). A clearer observation can be obtained from Fig. \ref{Fig:Comparison Between MACNP and VPMSCNP}(d). That is, MACNP$^{VP}$ obtains better $f_{avg}$ values on fifteen instances, worse $f_{avg}$ values only on two instances (i.e., \textit{USAir} and \textit{Ham5000}), and equal $f_{avg}$ values on the 9 remaining instances. These observations show that the state-of-the-art MACNP algorithm can also definitively benefit from the strategic population sizing mechanism proposed in this work.

\subsubsection{Comparisons with state-of-the-art algorithms}
\label{SubSubSec:Comparisons With State-of-the-Art Algorithms}

We report now a comparative study with respect to three recent state-of-the-art CNP algorithms, including CAN1 \cite{Pullan2015}, FastCNP \cite{Zhou2017b} and MACNP \cite{Zhou2019}. To the best of our knowledge, the best-known results available in the literature have been achieved by these three algorithms. Detailed comparative results between our algorithms (i.e., VPMS$_{CNP}$ and MACNP$^{VP}$) and the reference algorithms are shown in Table \ref{Tab:Comparative Performance Under the Time Limit 3600 Seconds}.

\begin{table*}[!htbp]
\caption{Comparisons of our algorithms with state-of-the-art algorithms under $t_{max} = 3600$ seconds.}
\label{Tab:Comparative Performance Under the Time Limit 3600 Seconds}
\begin{center}
\begin{scriptsize}
\begin{threeparttable}
\begin{tabular}{lrrrrrrrrrrr}
\toprule[0.75pt]
\multicolumn{2}{c}{} & \multicolumn{2}{c}{CAN1\cite{Pullan2015}} & \multicolumn{2}{c}{FastCNP\cite{Zhou2017b}} & \multicolumn{2}{c}{MACNP\cite{Zhou2019}} & \multicolumn{2}{c}{\textbf{MACNP$^{VP}$}} & \multicolumn{2}{c}{\textbf{VPMS$_{CNP}$}}\\
\cmidrule[0.5pt]{3-4} \cmidrule[0.5pt]{5-6} \cmidrule[0.5pt]{7-8} \cmidrule[0.5pt]{9-10} \cmidrule[0.5pt]{11-12}
Instance & $f_{bkv}$ & $f_{best}$ & $f_{avg}$ & $f_{best}$ & $f_{avg}$ & $f_{best}$ & $f_{avg}$ & $f_{best}$ & $f_{avg}$ & $f_{best}$ & $f_{avg}$ \\
\midrule[0.5pt]
BA500&195&\textbf{195}&\textbf{195.0}&\textbf{195}&\textbf{195.0}&\textbf{195}&\textbf{195.0}&\textbf{195}&\textbf{195.0}&\textbf{195}&\textbf{195.0}\\
BA1000&558&\textbf{558}&558.7&\textbf{558}&\textbf{558.0}&\textbf{558}&\textbf{558.0}&\textbf{558}&\textbf{558.0}&\textbf{558}&\textbf{558.0}\\
BA2500&3704&\textbf{3704}&\textbf{3704.0}&\textbf{3704}&3710.6&\textbf{3704}&\textbf{3704.0}&\textbf{3704}&\textbf{3704.0}&\textbf{3704}&\textbf{3704.0}\\
BA5000&10196&\textbf{10196}&\textbf{10196.0}&\textbf{10196}&10201.4&\textbf{10196}&\textbf{10196.0}&\textbf{10196}&\textbf{10196.0}&\textbf{10196}&\textbf{10196.0}\\
ER235&295&\textbf{295}&\textbf{295.0}&\textbf{295}&\textbf{295.0}&\textbf{295}&\textbf{295.0}&\textbf{295}&\textbf{295.0}&\textbf{295}&\textbf{295.0}\\
ER466&1524&\textbf{1524}&\textbf{1524.0}&\textbf{1524}&\textbf{1524.0}&\textbf{1524}&\textbf{1524.0}&\textbf{1524}&\textbf{1524.0}&\textbf{1524}&\textbf{1524.0}\\
ER941&5012&5114&5177.4&\textbf{5012}&\textbf{5013.3}&\textbf{5012}&5014.1&\textbf{5012}&5015.9&\textbf{5012}&5026.5\\
ER2344&902498&996411&1008876.4&953437&979729.2&\textbf{902498}&\textbf{922339.5}&912205&929024.1&904113&933943.7\\
FF250&194&\textbf{194}&\textbf{194.0}&\textbf{194}&\textbf{194.0}&\textbf{194}&\textbf{194.0}&\textbf{194}&\textbf{194.0}&\textbf{194}&\textbf{194.0}\\
FF500&257&263&265.0&\textbf{257}&258.4&\textbf{257}&\textbf{257.0}&\textbf{257}&\textbf{257.0}&\textbf{257}&\textbf{257.0}\\
FF1000&1260&1262&1264.2&\textbf{1260}&1260.8&\textbf{1260}&\textbf{1260.0}&\textbf{1260}&\textbf{1260.0}&\textbf{1260}&\textbf{1260.0}\\
FF2000&4545&4548&4549.4&4546&4558.3&\textbf{4545}&4545.7&\textbf{4545}&\textbf{4545.0}&\textbf{4545}&\textbf{4545.0}\\
WS250&3083&3415&3702.8&3085&3196.4&\textbf{3083}&3089.4&\textbf{3083}&3087.5&\textbf{3083}&\textbf{3083.1}\\
WS500&2072&2085&2098.7&\textbf{2072}&2083.3&\textbf{2072}&2082.6&\textbf{2072}&\textbf{2082.1}&\textbf{2072}&2083.1\\
WS1000&109807&141759&161488.0&123602&127493.4&\textbf{109807}&\textbf{123682.6}&123253&135187.8&119444&134475.5\\
WS1500&13098&13498&13902.5&13158&13255.7&\textbf{13098}&13255.1&\textbf{13098}&13175.7&\textbf{13098}&\textbf{13161.5}\\
\midrule[0.5pt]
Bovine&268&\textbf{268}&\textbf{268.0}&\textbf{268}&\textbf{268.0}&\textbf{268}&\textbf{268.0}&\textbf{268}&\textbf{268.0}&\textbf{268}&\textbf{268.0}\\
Circuit&2099&\textbf{2099}&\textbf{2099.0}&\textbf{2099}&\textbf{2099.0}&\textbf{2099}&\textbf{2099.0}&\textbf{2099}&\textbf{2099.0}&\textbf{2099}&\textbf{2099.0}\\
E.coli&806&\textbf{806}&\textbf{806.0}&\textbf{806}&\textbf{806.0}&\textbf{806}&\textbf{806.0}&\textbf{806}&\textbf{806.0}&\textbf{806}&\textbf{806.0}\\
USAir97&4336&\textbf{4336}&\textbf{4336.0}&\textbf{4336}&\textbf{4336.0}&\textbf{4336}&\textbf{4336.0}&\textbf{4336}&5275.0&\textbf{4336}&5075.6\\
HumanDisea&1115&\textbf{1115}&\textbf{1115.0}&\textbf{1115}&\textbf{1115.0}&\textbf{1115}&\textbf{1115.0}&\textbf{1115}&\textbf{1115.0}&\textbf{1115}&\textbf{1115.0}\\
Treni\_Roma&918&\textbf{918}&\textbf{918.0}&\textbf{918}&\textbf{918.0}&\textbf{918}&\textbf{918.0}&\textbf{918}&\textbf{918.0}&\textbf{918}&\textbf{918.0}\\
EU\_flights&348268&\textbf{348268}&\textbf{348347.0}&\textbf{348268}&348697.7&\textbf{348268}&351657.0&\textbf{348268}&349265.6&\textbf{348268}&349265.6\\
openflights&26842&29300&29815.3&28834&29014.4&26842&28704.3&26842&27792.3&\textbf{26785}\tnote{$\star$}&\textbf{27327.0}\\
yeast&1412&1413&1416.3&\textbf{1412}&\textbf{1412.0}&\textbf{1412}&\textbf{1412.0}&\textbf{1412}&\textbf{1412.0}&\textbf{1412}&\textbf{1412.0}\\
H1000&306349&314152&317805.7&314964&316814.8&\textbf{306349}&310626.5&306353&\textbf{310081.3}&307117&311169.4\\
H2000&1243859&1275968&1292400.4&1275204&1285629.1&1243859&1263495.6&\textbf{1242999}\tnote{$\star$}&\textbf{1251826.9}&1247652&1256573.8\\
H3000a&2844393&2911369&2927312.0&2885588&2906965.5&2844393&2884781.7&2842072\tnote{$\star$}&\textbf{2855005.3}&\textbf{2840941}\tnote{$\star$}&2859284.4\\
H3000b&2841270&2907643&2927330.5&2876585&2902893.9&2841270&2885087.0&\textbf{2839018}\tnote{$\star$}&\textbf{2847010.7}&2839893\tnote{$\star$}&2860810.9\\
H3000c&2838429&2885836&2917685.8&2876026&2898879.3&2838429&2869348.5&2834802\tnote{$\star$}&\textbf{2843661.7}&\textbf{2832073}\tnote{$\star$}&2844324.3\\
H3000d&2831311&2906121&2929569.2&2894492&2907485.4&2831311&2892562.7&\textbf{2827859}\tnote{$\star$}&\textbf{2846261.0}&2830291\tnote{$\star$}&2857201.4\\
H3000e&2847909&2903845&2931806.8&2890861&2911409.3&2847909&2887525.7&\textbf{2846412}\tnote{$\star$}&\textbf{2855333.6}&2846731\tnote{$\star$}&2867000.6\\
H4000&5044357&5194592&5233954.5&5167043&5190883.7&\textbf{5044357}&5137528.3&5077298&\textbf{5125589.3}&5082521&5141804.3\\
H5000&7972525&8142430&8212165.9&8080473&8132896.2&\textbf{7972525}&\textbf{8094812.6}&8012229&8120955.9&8011565&8151850.1\\
powergr&15862&16158&16222.1&15982&16033.5&\textbf{15862}&15901.5&15870&\textbf{15897.1}&15873&15909.2\\
Oclinks&611326&611326&614858.5&611344&616783.0&612303&614544.0&611280\tnote{$\star$}&614364.0&\textbf{611254}\tnote{$\star$}&\textbf{614296.3}\\
faceboo&420334&701073&742688.0&692799&765609.8&\textbf{643162}&\textbf{739436.6}&687604&760335.1&691232&780429.1\\
grqc&13596&15522&15715.7&13616&13634.8&13596&13629.2&\textbf{13592}\tnote{$\star$}&\textbf{13611.4}&13603&13615.5\\
hepth&106397&130256&188753.7&108217&109889.5&\textbf{106397}&\textbf{109655.6}&106778&108961.1&107939&110158.4\\
hepph&6156536&9771610&10377853.2&\textbf{6392653}&\textbf{7055773.8}&8628687&9370215.3&7465746&8128758.7&7883063&8689170.1\\
astroph&53963375&59029312&60313225.8&\textbf{55424575}&\textbf{57231348.7}&62068966&62547898.1&57411990&59897908.4&58322396&59563941.1\\
condmat&2298596&13420836&14823254.9&\textbf{4086629}&\textbf{5806623.8}&9454361&10061807.8&6438018&7407961.4&6843993&7813436.7\\
%\midrule[0.5pt]
%better/equal/worse & &\textbf{0/14/28}&\textbf{3/11/28} &\textbf{3/19/20}&\textbf{7/11/24} &8/23/11&\textbf{6/15/21} &$-$&$-$ &7/22/13&\textbf{6/18/18}\\
%\midrule[0.5pt]
%better/equal/worse & &\textbf{0/14/28}&\textbf{3/11/28} &\textbf{3/19/20}&\textbf{9/11/22} &10/22/10&11/15/16 &13/22/7&18/17/7 &$-$&$-$\\
\bottomrule[0.75pt]
\end{tabular}
\begin{tablenotes}
    \item [$\star$] Improved best upper bounds.
\end{tablenotes}
\end{threeparttable}
\end{scriptsize}
\end{center}
\end{table*}

Table \ref{Tab:Comparative Performance Under the Time Limit 3600 Seconds} shows that both VPMS$_{CNP}$ and MACNP$^{VP}$ achieve highly competitive performances compared to the reference algorithms. Under the time limit $t_{max} = 3600$ seconds, these algorithms attain 9 new upper bounds and match 22 known best upper bounds. At a significant level of 0.05, VPMS$_{CNP}$ is significantly better than CAN1 (i.e., $35 > CV^{42}_{0.05} = 27$) and FastCNP (i.e., $29.5 > CV^{42}_{0.05} = 27$) in terms of $f_{best}$. For the $f_{avg}$ indicator, both VPMS$_{CNP}$ and MACNP$^{VP}$ once again significantly outperform CAN1 and FastCNP. Compared to MACNP, VPMS$_{CNP}$ performs better, leading to better $f_{best}$ values on 11 instances, and equal $f_{best}$ values on 23 instances, but the performance difference is statistically marginal (i.e., $22.5 < CV^{42}_{0.05} = 27$). For the $f_{avg}$ indicator, MACNP$^{VP}$ is significantly better than MACNP (i.e., $28.5 > CV^{42}_{0.05} = 27$) and VPMS$_{CNP}$ ($27 \geqslant CV^{42}_{0.05} = 27$). These findings show that memetic algorithms using a variable population compete favorably with the state-of-the-art CNP algorithms.

\begin{table*}[!htbp]
\caption{Comparisons of the variable population memetic algorithms (MACNP$^{VP}$ and VPMS$_{CNP}$) with the fixed population memetic MACNP algorithm under $t_{max} = 7200$ seconds.}
\label{Tab:Comparisons Between VPMSCNP Algorithms and MACNP Algorithm Under the Time Limit 7200 Seconds}
\begin{center}
\begin{scriptsize}
\begin{threeparttable}
\begin{tabular}{lrrrrrrr}
\toprule[0.75pt]
\multicolumn{2}{c}{} & \multicolumn{2}{c}{MACNP\tnote{$\ast$}} & \multicolumn{2}{c}{\textbf{MACNP$^{VP}$}} & \multicolumn{2}{c}{\textbf{VPMS$_{CNP}$}}\\
\cmidrule[0.5pt]{3-4} \cmidrule[0.5pt]{5-6} \cmidrule[0.5pt]{7-8}
Instance & $f_{bkv}$ & $f_{best}$ & $f_{avg}$ & $f_{best}$ & $f_{avg}$ & $f_{best}$ & $f_{avg}$  \\
\midrule[0.5pt]
BA500&195&\textbf{195}&\textbf{195.0} &\textbf{195}&\textbf{195.0} &\textbf{195}&\textbf{195.0}\\
BA1000&558&\textbf{558}&558.1&\textbf{558}&\textbf{558.0} &\textbf{558}&\textbf{558.0}\\
BA2500&3704&\textbf{3704}&\textbf{3704.0} &\textbf{3704}&\textbf{3704.0} &\textbf{3704}&\textbf{3704.0}\\
BA5000&10196&\textbf{10196}&\textbf{10196.0} &\textbf{10196}&\textbf{10196.0} &\textbf{10196}&\textbf{10196.0}\\
ER235&295&\textbf{295}&\textbf{295.0} &\textbf{295}&\textbf{295.0} &\textbf{295}&\textbf{295.0}\\
ER466&1524&\textbf{1524}&\textbf{1524.0} &\textbf{1524}&\textbf{1524.0} &\textbf{1524}&\textbf{1524.0}\\
ER941&5012&\textbf{5012}&5016.1&\textbf{5012}&\textbf{5015.9} &\textbf{5012}&5026.5\\
ER2344&902498&911274&929358.3&912205&\textbf{929024.1} &\textbf{904113}&933943.7\\
FF250&194&\textbf{194}&\textbf{194.0} &\textbf{194}&\textbf{194.0} &\textbf{194}&\textbf{194.0}\\
FF500&257&\textbf{257}&257.3&\textbf{257}&\textbf{257.0} &\textbf{257}&\textbf{257.0}\\
FF1000&1260&\textbf{1260}&1262.6&\textbf{1260}&\textbf{1260.0} &\textbf{1260}&\textbf{1260.0}\\
FF2000&4545&\textbf{4545}&4547.6&\textbf{4545}&\textbf{4545.0} &\textbf{4545}&\textbf{4545.0}\\
WS250&3083&\textbf{3083}&3083.2&\textbf{3083}&3087.5&\textbf{3083}&\textbf{3083.1}\\
WS500&2072&\textbf{2072}&2086.1&\textbf{2072}&\textbf{2082.1} &\textbf{2072}&2083.1\\
WS1000&109807&\textbf{110342}&\textbf{125548.4} &123253&135187.8&119444&134475.5\\
WS1500&13098&13150&13339.1&\textbf{13098}&13175.7&\textbf{13098}&\textbf{13161.5}\\
\midrule[0.5pt]
Bovine&268&\textbf{268}&\textbf{268.0} &\textbf{268}&\textbf{268.0} &\textbf{268}&\textbf{268.0}\\
Circuit&2099&\textbf{2099}&\textbf{2099.0} &\textbf{2099}&\textbf{2099.0} &\textbf{2099}&\textbf{2099.0}\\
Ecoli.txt&806&\textbf{806}&\textbf{806.0} &\textbf{806}&\textbf{806.0} &\textbf{806}&\textbf{806.0} \\
USAir97&4336&\textbf{4336}&\textbf{4343.1} &\textbf{4336}&5151.5&\textbf{4336}&5060.3 \\
humanDisea&1115&\textbf{1115}&\textbf{1115.0} &\textbf{1115}&\textbf{1115.0} &\textbf{1115}&\textbf{1115.0} \\
Treni\_Roma&918&\textbf{918}&\textbf{918.0} &\textbf{918}&\textbf{918.0} &\textbf{918}&\textbf{918.0} \\
EU\_flights&348268&\textbf{348268}&351573.9&\textbf{348268}&349016.2&\textbf{348268}&\textbf{348434.3} \\
openflights&26842&26842&28724.9&26842&27821.1&\textbf{26783}\tnote{$\star$}&\textbf{26919.0} \\
yeast1&1412&\textbf{1412}&1412.6&\textbf{1412}&\textbf{1412.0} &\textbf{1412}&\textbf{1412.0} \\
Ham1000&306349&306353&310254.2&\textbf{306349}&310348.5&\textbf{306349}&\textbf{309912.0} \\
Ham2000&1243859&1243810\tnote{$\star$}&1255525.9&\textbf{1242739}\tnote{$\star$}&\textbf{1249217.5} &1242792\tnote{$\star$}&1251189.7 \\
Ham3000a&2844393&2841893\tnote{$\star$}&2851070.2&2841487\tnote{$\star$}\tnote{$\star$}&\textbf{2845235.8} &\textbf{2840690}\tnote{$\star$}&2847291.7 \\
Ham3000b&2841270&2839435\tnote{$\star$}&2845280.4&2839098\tnote{$\star$}&\textbf{2841822.5} &\textbf{2837584}\tnote{$\star$}&2843768.2 \\
Ham3000c&2838429&2836103\tnote{$\star$}&2841923&\textbf{2835369}\tnote{$\star$}&\textbf{2837858.0} &2835860\tnote{$\star$}&2839192.3 \\
Ham3000d&2831311&2829328\tnote{$\star$}&2839602.4&\textbf{2828492}\tnote{$\star$}&\textbf{2834729.6} &2829102\tnote{$\star$}&2841551.0 \\
Ham3000e&2847909&2844979\tnote{$\star$}&2858484.1&2845437\tnote{$\star$}&2850598.1&\textbf{2843000}\tnote{$\star$}&\textbf{2847442.4} \\
Ham4000&5044357&5042395\tnote{$\star$}&5105351.2&5045783&\textbf{5089596.9} &\textbf{5038611}\tnote{$\star$}&5091745.6 \\
Ham5000&7972525&\textbf{7964765}\tnote{$\star$}&8060826&7969299\tnote{$\star$}&\textbf{8039418.4} &7969845\tnote{$\star$}&8042058.9 \\
powergrid&15862&15897&15943.7&\textbf{15865}&\textbf{15882.7} &15868&15886.1 \\
Oclinks&611326&612328&614732.8&\textbf{611253}\tnote{$\star$}&\textbf{613861.5} &611260\tnote{$\star$}&614220.9 \\
facebook&420334&680936&783374.6&\textbf{630564}&\textbf{732633.6} &669910&738856.5 \\
grqc&13596&13601&13644&\textbf{13591}\tnote{$\star$}&\textbf{13598.4} &13592\tnote{$\star$}&13602.4 \\
hepth&106397&106926&108238.5&\textbf{106276}\tnote{$\star$}&\textbf{108079.9} &106792&108673.4\\
hepph&6156536&\textbf{6155877}\tnote{$\star$}&\textbf{6991782.6} &7087968&7724431.6&7211646&7960148.5 \\
astroph&53963375&58941340&60665177.4&\textbf{55800209}&\textbf{56920216.6} &56229708&57421239.1\\
condmat&2298596&\textbf{5205685}&6580912.8&5393192&\textbf{6403204.6}&6057949&6593803.2 \\
%\midrule[0.5pt]
%better/equal/worse& &7/22/13&\textbf{5/11/26} &$-$&$-$ &7/23/12&8/16/18\\
%\midrule[0.5pt]
%better/equal/worse& &\textbf{4/21/17}&\textbf{7/11/24} &12/23/7&18/16/8 &$-$&$-$\\
\bottomrule[0.75pt]
\end{tabular}
\begin{tablenotes}
    \item[$\ast$] The results are obtained by re-running MACNP \cite{Zhou2019} with $t_{max} = 7200$ seconds.
    \item [$\star$] Improved best upper bounds.
\end{tablenotes}
\end{threeparttable}
\end{scriptsize}
\end{center}
\end{table*}

To further study the behavior of the two memetic algorithms with a variable population (i.e., VPMS$_{CNP}$ and MACNP$^{VP}$) under long time limits, we also compared them against the MACNP algorithm, which uses a fixed population, with a relaxed time limit $t_{max} = 7200$ seconds. The comparative results are summarized in Table \ref{Tab:Comparisons Between VPMSCNP Algorithms and MACNP Algorithm Under the Time Limit 7200 Seconds}. We observe that both VPMS$_{CNP}$ and MACNP$^{VP}$ achieve still better results. For the 42 benchmark instances, VPMS$_{CNP}$ and MACNP$^{VP}$ find 13 new upper bounds and reach 23 best-known solutions. At a significant level of 0.05, MACNP$^{VP}$ performs significantly better than MACNP (i.e., $31.5 > CV^{42}_{0.05} = 27$) in terms of the $f_{avg}$ indicator. For the $f_{best}$ indicator, MACNP$^{VP}$ performs marginally better than MACNP with a better result on 13 instances, an equal result on 22 instances and a worse result on 7 instances (i.e., $24 < CV^{42}_{0.05} = 27$). Remarkably, VPMS$_{CNP}$ significantly outperforms MACNP both in terms of $f_{best}$ (i.e., $27.5 > CV^{42}_{0.05} = 27$) and $f_{avg}$ (i.e., $29.5 > CV^{42}_{0.05} = 27$). These observations further demonstrate the relevance of the strategic population sizing strategy for enhancing memetic algorithms.

\section{Conclusion and future work}
\label{Sec:Conclusions and Future Work}

In this work, we presented the variable population memetic search (VPMS) framework where a strategic population sizing mechanism is introduced into memetic algorithms to dynamically adjust the population size during the evolutionary search. Unlike the conventional memetic search framework, our VPMS approach starts its search from a population of only two solutions, and dynamically increases or decreases the population size under specific rules. By strategically varying the population size, the memetic algorithm is able to adapt the population diversity during the search and thus favors a continuing balancing between exploitation and exploration. To demonstrate the effectiveness of the proposed VPMS approach, we presented a case study by applying VPMS to solve the challenging critical node problem where a diversified late acceptance search procedure for CNP was designed as the local improvement component of the VPMS algorithm.

Extensive computational studies on two sets of 42 (both synthetic and real-world) benchmark instances in the literature showed that our approach with a variable population competes very favorably with the state-of-the-art CNP algorithms, and remarkably discovers new upper bounds for 13 instances. This study also confirmed the benefit of the strategic population sizing mechanism as a general technique to improve the performance of the classical memetic search. For future work, one  research perspective is to investigate the application of the VPMS approach to solve other combinatorial problems. Another interesting research is to determine the population size according to more elaborated rules that can rely on refined information acquired from machine learning techniques.

%% use section* for acknowledgment
%\section*{Acknowledgment}
%
%We would like to thank the referees for their useful comments and suggestions.

\ifCLASSOPTIONcaptionsoff
  \newpage
\fi

\bibliography{IEEEexample}

\begin{thebibliography}{1}

\bibitem{Addis2013}
B.~Addis, M.~D. Summa, and A.~Grosso,
``Identifying critical nodes in undirected graphs: Complexity results and polynomial algorithms for the case of bounded treewidth,''
\emph{Discrete Applied Mathematics}, vol.~161, no.~16-17, pp. 2349--2360, 2013.

\bibitem{Addis2016}
B.~Addis, R.~Aringhieri, A.~Grosso, and P.~Hosteins,
``Hybrid constructive heuristics for the critical node problem,''
\emph{Annals of Operations Research}, vol.~238, no.~1, pp. 1--13, 2016.

\bibitem{Arabas1994}
J.~Arabas, Z.~Michalewicz, and J.~Mulawka,
``{GAVaPS}-a genetic algorithm with varying population size,''
in \emph{Proceedings of the First IEEE Conference on Evolutionary Computation. IEEE World Congress on Computational Intelligence}. IEEE, 1994, pp. 73--78.

\bibitem{Aringhieri2016a}
R.~Aringhieri, A.~Grosso, P.~Hosteins, and R.~Scatamacchia,
``A general evolutionary framework for different classes of critical node problems,''
\emph{Engineering Applications of Artificial Intelligence}, vol.~55, no.~C, pp. 128--145, 2016.

\bibitem{Aringhieri2016b}
R.~Aringhieri, A.~Grosso, P.~Hosteins, and R.~Scatamacchia,
``Local search metaheuristics for the critical node problem,''
\emph{Networks}, vol.~67, no.~3, pp. 209--221, 2016.

\bibitem{Arulselvan2009}
A.~Arulselvan, C.~W. Commander, L.~Elefteriadou, and P.~M. Pardalos,
``Detecting critical nodes in sparse graphs,''
\emph{Computers \& Operations Research}, vol.~36, no.~7, pp. 2193--2200, 2009.

\bibitem{Benlic2015}
U.~Benlic and J.-K. Hao,
``Memetic search for the quadratic assignment problem,''
\emph{Expert Systems with Applications}, vol.~42, no.~1, pp. 584--595, 2015.

\bibitem{Brest2012}
J.~Brest, A.~Zamuda, I.~Fister, M.~S. Mau{\v{c}}ec \emph{et~al.},
``Self-adaptive differential evolution algorithm with a small and varying population size,''
in \emph{2012 IEEE Congress on Evolutionary Computation}. IEEE, 2012, pp. 1--8.

\bibitem{Burke2017}
E.~K. Burke and Y.~Bykov,
``The late acceptance hill-climbing heuristic,''
\emph{European Journal of Operational Research}, vol.~258, no.~1, pp. 70--78, 2017.

\bibitem{Chen2011}
X.~Chen, Y.-S. Ong, M.-H. Lim, and K.~C. Tan,
``A multi-facet survey on memetic computation,''
\emph{IEEE Transactions on Evolutionary Computation}, vol.~15, no.~5, pp. 591--607, 2011.

\bibitem{Cormen2009}
T.~H. Cormen, C.~E. Leiserson, R.~L. Rivest, and C.~Stein,
\emph{Introduction to Algorithms, 3rd Edition}. MIT Press, 2009.

\bibitem{Demvsar2006}
J.~Dem{\v{s}}ar,
``Statistical comparisons of classifiers over multiple data sets,''
\emph{Journal of Machine Learning Research}, vol.~7, pp. 1--30, 2006.

\bibitem{Dinh2011}
T.~N. Dinh and M.~T. Thai,
``Precise structural vulnerability assessment via mathematical programming,''
in \emph{Military Communications Conference, 2011-MILCOM 2011}. IEEE, 2011, pp. 1351--1356.

\bibitem{Eiben2004}
A.~E. Eiben, E.~Marchiori, and V.~Valko,
``Evolutionary algorithms with on-the-fly population size adjustment,''
in \emph{International Conference on Parallel Problem Solving from Nature}. Springer, 2004, pp. 41--50.

\bibitem{Feng2014}
L.~Feng, Y.-S. Ong, M.-H. Lim, and I.~W. Tsang,
``Memetic search with interdomain learning: A realization between {CVRP} and {CARP},''
\emph{IEEE Transactions on Evolutionary Computation}, vol.~19, no.~5, pp. 644--658, 2014.

\bibitem{Fernandez2003}
F.~Fern{\'a}ndez, M.~Tomassini, and L.~Vanneschi,
``An empirical study of multipopulation genetic programming,''
\emph{Genetic Programming and Evolvable Machines}, vol.~4, no.~1, pp. 21--51, 2003.

\bibitem{Guan2017}
Y.~Guan, L.~Yang, and W.~Sheng,
``Population control in evolutionary algorithms: Review and Comparison,''
\emph{Bio-inspired Computing: Theories and Applications}, pp. 161--174, 2017.

\bibitem{Hart2004}
W.~E. Hart, N.~Krasnogor, and J.~E. Smith,
``Recent Advances in Memetic Algorithms,''
\emph{Studies in Fuzziness and Soft Computing}, vol.~166, Springer Science \& Business Media, 2004.


\bibitem{Hoos2004}
H.~H. Hoos and T.~St{\"u}tzle,
\emph{Stochastic Local Search: Foundations and Applications}. Elsevier, 2004.

\bibitem{Hopcroft1973}
J.~Hopcroft and R.~Tarjan,
``Algorithm 447: efficient algorithms for graph manipulation,''
\emph{Communications of the ACM}, vol.~16, no.~6, pp. 372--378, 1973.

\bibitem{JinHao2019}
Y.~Jin, and J-K.~Hao, ``Solving the latin square completion problem by memetic graph coloring,'' \emph{IEEE Transactions on Evolutionary Computation}, online since February 2019, DOI: 10.1109/TEVC.2019.2899053.

\bibitem{Karapetyan2011}
D.~Karapetyan and G.~Gutin,
``A new approach to population sizing for memetic algorithms: a case study for the multidimensional assignment problem,''
\emph{Evolutionary Computation}, vol.~19, no.~3, pp. 345--371, 2011.

\bibitem{Koumousis2006}
V.~K. Koumousis and C.~P. Katsaras,
``A saw-tooth genetic algorithm combining the effects of variable population size and reinitialization to enhance performance,'' \emph{IEEE Transactions on Evolutionary Computation}, vol.~10, no.~1, pp. 19--28, 2006.

\bibitem{Krasnogor2005}
N.~Krasnogor and J.~Smith,
``A tutorial for competent memetic algorithms: model, taxonomy, and design issues,''
\emph{IEEE Transactions on Evolutionary Computation}, vol.~9, no.~5, pp. 474--488, 2005.

\bibitem{Lalou2018}
M.~Lalou, M.~A. Tahraoui, and H.~Kheddouci,
``The critical node detection problem in networks: A survey,''
\emph{Computer Science Review}, vol.~28, pp. 92--117, 2018.

\bibitem{Martins2018}
D.~Martins, G.~M. Viana, I.~Rosseti, S.~L. Martins, and A.~Plastino,
``Making a state-of-the-art heuristic faster with data mining,''
\emph{Annals of Operations Research}, vol.~263, pp. 141-162, 2018.

\bibitem{Moscato&Cotta2003}
P. Moscato and C. Cotta,
``A gentle introduction to memetic algorithms,''
In \emph{Handbook of Metaheuristics}, Kluwer, Norwell, Massachusetts, USA, pp. 105--144, 2003.

\bibitem{Namazi2018}
M.~Namazi, C.~Sanderson, M.~A.~H. Newton, M.~M.~A. Polash, and A.~Sattar,
``Diversified late acceptance search,''
in \emph{{AI} 2018: Advances in Artificial Intelligence - 31st Australasian Joint Conference, Wellington, New Zealand, December 11-14, 2018, Proceedings}, pp. 299--311, 2018.

\bibitem{Neri2012}
F.~Neri and C.~Cotta,
``Memetic algorithms and memetic computing optimization: A literature review,''
\emph{Swarm and Evolutionary Computation}, vol.~2, pp. 1--14, 2012.

\bibitem{Pavai2016}
G.~Pavai and T.~V. Geetha,
``A survey on crossover operators,''
\emph{ACM Computing Surveys}, vol.~49, no.~4, pp. 72:1--72:43, 2016.

\bibitem{Porumbeletal2010}
D.~C. Porumbel, J.~K. Hao, and P.~Kuntz,
``An evolutionary approach with diversity guarantee and well-informed grouping recombination for graph coloring,'' \emph{Computers \& Operations Research}, vol.~37, no.~10, pp. 1822--1832, 2010.

\bibitem{Pullan2015}
W.~Pullan,
``Heuristic identification of critical nodes in sparse real-world graphs,''
\emph{Journal of Heuristics}, vol.~21, no.~5, pp. 577--598, 2015.

\bibitem{Segura2017}
C.~Segura, A.~H. Aguirre, F.~Luna and E.~Alba,
``Improving diversity in evolutionary algorithms:New best solutions for frequency assignment,''
\emph{IEEE Transactions on Evolutionary Computation}, vol.~21, no.~4, pp. 539--553, 2017.

\bibitem{Sorensen2006}
K.~S{\"{o}}rensen and M.~Sevaux,
``{MA} $\mid$ {PM}: memetic algorithms with population management,''
\emph{Computers \& Operations Research}, vol.~33, no.~5, pp. 1214--1225, 2006.

\bibitem{Summa2011}
M.~D. Summa, A.~Grosso, and M.~Locatelli,
``Complexity of the critical node problem over trees,''
\emph{Computers \& Operations Research}, vol.~38, no.~12, pp. 1766--1774, 2011.

\bibitem{Summa2012}
M.~D. Summa, A.~Grosso, and M.~Locatelli,
``Branch and cut algorithms for detecting critical nodes in undirected graphs,''
\emph{Computational Optimization and Applications}, vol.~53, no.~3, pp. 649--680, 2012.

\bibitem{Tang2009}
K.~Tang, Y.~Mei, and X.~Yao,
``Memetic algorithm with extended neighborhood search for capacitated arc routing problems,''
\emph{IEEE Transactions on Evolutionary Computation}, vol.~13, no.~5, pp. 1151-1166, 2009.

\bibitem{Tirronen2009}
V.~Tirronen and F.~Neri,
``Differential evolution with fitness diversity self-adaptation,''
in \emph{Nature-Inspired Algorithms for Optimisation}. Springer, 2009, pp. 199--234.

\bibitem{Ventresca2012}
M.~Ventresca,
``Global search algorithms using a combinatorial unranking-based problem representation for the critical node detection problem,''
\emph{Computers \& Operations Research}, vol.~39, no.~11, pp. 2763--2775, 2012.

\bibitem{Ventresca2014}
M.~Ventresca and D.~Aleman,
``A fast greedy algorithm for the critical node detection problem,''
in \emph{International Conference on Combinatorial Optimization and Applications}. Springer, 2014, pp. 603--612.

\bibitem{Veremyev2014}
A.~Veremyev, V.~Boginski, and E.~L. Pasiliao,
``Exact identification of critical nodes in sparse networks via new compact formulations,''
\emph{Optimization Letters}, vol.~8, no.~4, pp. 1245--1259, 2014.

\bibitem{Wang2019}
S.~Wang, J.~Liu, and Y.~Jin,
``Finding influential nodes in multiplex networks using a memetic algorithm,''
\emph{IEEE Transactions on Cybernetics}, DOI: 10.1109/TCYB.2019.2917059, 2019.

\bibitem{Weise2016}
T.~Weise, Y.~Wu, R.~Chiong, K.~Tang, and J.~L{\"a}ssig,
``Global versus local search: the impact of population sizes on evolutionary algorithm performance,''
\emph{Journal of Global Optimization}, vol.~66, no.~3, pp. 511--534, 2016.

\bibitem{Wu2019}
J.~Wu, X.~Shen, and K.~Jiao,
``Game-based memetic algorithm to the vertex cover of networks,''
\emph{IEEE Transactions on Cybernetics}, vol.~49, no.~3, pp. 974--988, 2019.

\bibitem{Zhou2017a}
Y.~Zhou, J.-K. Hao, and B.~Duval, ``Opposition-based memetic search for the maximum diversity problem,''
\emph{{IEEE} Transactions on Evolutionary Computation}, vol.~21, no.~5, pp. 731--745, 2017.

\bibitem{Zhou2017b}
Y.~Zhou and J.-K. Hao,
``A fast heuristic algorithm for the critical node problem,''
in \emph{Proceedings of the Genetic and Evolutionary Computation Conference Companion}. ACM, 2017, pp. 121--122.

\bibitem{Zhou2019}
Y.~Zhou, J.-K. Hao, and G.~Fred,
``Memetic search for identifying critical nodes in sparse graphs,''
\emph{{IEEE} Transactions on Cybernetics}, vol.~49, no.~10, pp. 3699--3712, Oct 2019.

\end{thebibliography}

\end{document}